\pdfoutput=1

\documentclass[11pt]{article}

\usepackage[final]{acl}
\usepackage{enumitem}
\usepackage{times}
\usepackage{latexsym}

\usepackage[T1]{fontenc}

\usepackage[utf8]{inputenc}

\usepackage{microtype}

\usepackage{inconsolata}

\usepackage{graphicx}

\usepackage{gb4e}
\usepackage{multirow}
\noautomath
\usepackage{subcaption}
\usepackage{tabularx}
\usepackage{float}
\usepackage{afterpage}
\usepackage{listings}
\usepackage{booktabs, tabularx, array} 
\usepackage[table]{xcolor}   
\usepackage[normalem]{ulem}
\usepackage{comment}

\title{\textsc{Correct-Detect}: Balancing Performance and Ambiguity Through the Lens of Coreference Resolution in LLMs}

\author{
  Amber Shore\textsuperscript{1} \quad
  Russell Scheinberg\textsuperscript{1} \quad
  Ameeta Agrawal\textsuperscript{1} \quad
  So Young Lee\textsuperscript{2} \\[4pt]
  \textsuperscript{1}\,Portland State University, USA \\
  \textsuperscript{2}\,Miami University, USA \\[4pt]
  \texttt{\{ashore, rschein2, ameeta\}@pdx.edu} \quad
  \texttt{soyoung.lee@miamioh.edu}
}

\begin{document}
\maketitle
\begin{abstract}
Large Language Models (LLMs) are intended to reflect human linguistic competencies. But humans have access to a broad and embodied context, which is key in detecting and resolving linguistic ambiguities, even in isolated text spans. A foundational case of semantic ambiguity is found in the task of coreference resolution: how is a pronoun related to an earlier person mention? This capability is implicit in nearly every downstream task, and the presence of ambiguity at this level can alter performance significantly. We show that LLMs can achieve good performance with minimal prompting in both coreference disambiguation and the detection of ambiguity in coreference, however, they cannot do both at the same time. We present the \textsc{Correct}-\textsc{Detect} trade-off: though models have both capabilities and deploy them implicitly, successful performance balancing these two abilities remains elusive. 
\end{abstract}

\section{Introduction}

Ambiguity resolution is fundamental to successful communication. Typically, context provides cues for resolving ambiguities \citep{bousquet2019useofcontext}, but when context is absent or insufficient, ambiguity resolution becomes more tenuous. 
Unlike the context-rich setting of human interaction, large language models (LLMs) operate with a significant contextual deficit: 
an LLM shares no social or physical context with its human user, and this lack of shared context means less common ground and fewer contextual cues are available to help language models resolve ambiguity (c.f. the open-domain paradox in \citet{skantze-dogruoz-2023-open}).
\begin{figure}[h]
\centering
\includegraphics[width=\columnwidth]{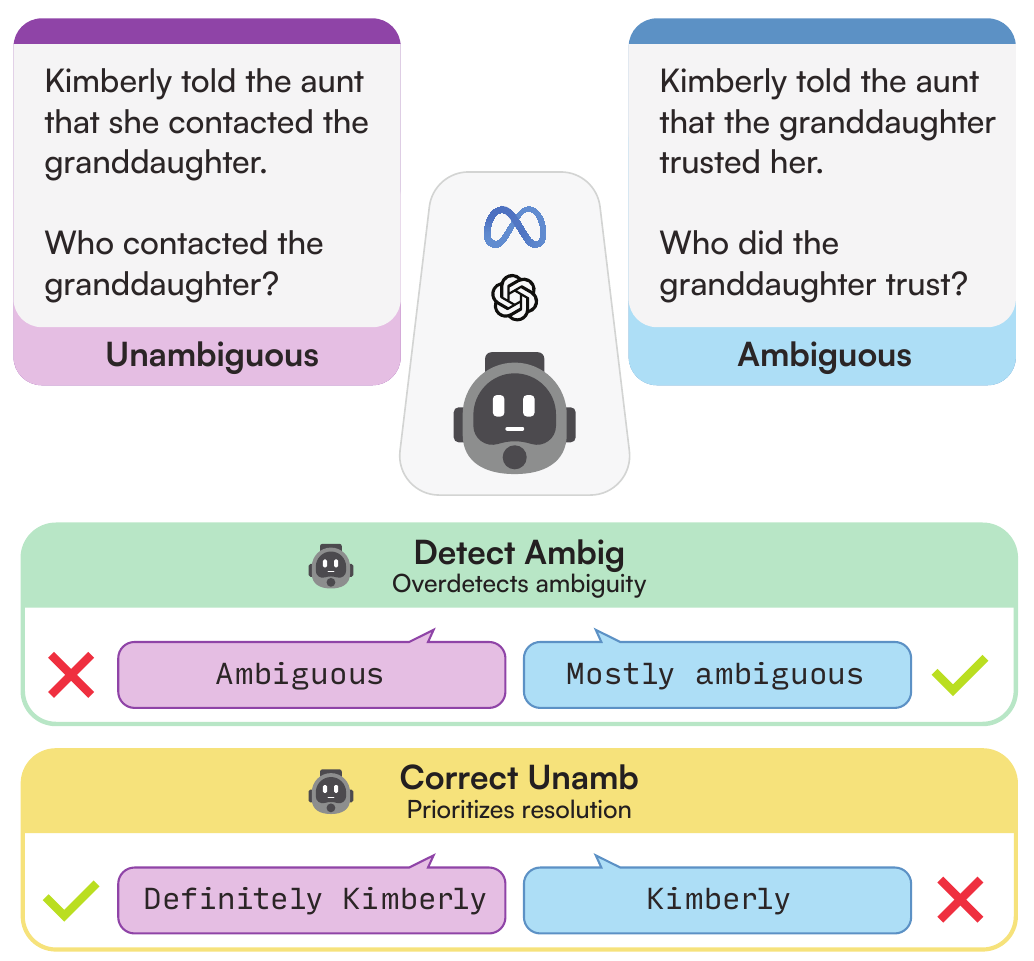}
\caption{
Our results suggest a trade-off between coreference resolution accuracy in unambiguous sentences (\textbf{Correct Unamb}) and reliable ambiguity detection in ambiguous sentences (\textbf{Detect Ambig}).
} \vspace{-0.5cm}
\label{fig:intro}
\end{figure}

Yet higher-level tasks such as summarization and question answering implicitly assume that LLMs can detect and resolve referential ambiguity.
For instance, take Winograd's classic sentence ``The city councilmen refused the demonstrators a permit because \textbf{they} feared/advocated violence'' \citep{WINOGRAD19721}: failure of LLMs to correctly identify the referent of \textit{they} could result in an inaccurate downstream response. Thus, LLMs need to be able to accurately interpret textual cues for disambiguation, \textit{and to recognize when textual cues are insufficient}. 
How well do they apply subtle semantic cues for disambiguation? Can they distinguish when ambiguity is resolvable and when it is not? {See Figure \ref{fig:intro} for a demonstration of these competing goals.}

To probe how LLMs handle ambiguity, we focus on coreference resolution -- a prototypical case of referential disambiguation. 
Coreference resolution is one of the ways implicit disambiguation shows up as a foundational task for language models. \citet{lampinen2024broaderspectrumincontextlearning} argue that it forms one of the ``potential roots'' of generalized in-context learning in LLMs.
Coreference resolution is the task of determining which text spans refer to the same entity, a necessary component of discourse parsing \citep{liu2023surveycorefresolution}.
For example, the pronoun ``she'' in Example (1) refers to ``Anna'', and since both expressions refer to the same entity, they are co-referents.
\begin{exe} \label{intro_ex1}
\item "Anna$_i$ looked out the window. She$_i$ saw that the rain had stopped."
\end{exe}
In some cases, the presence of multiple entities can lead to genuine ambiguity, as in Example (2):
\begin{exe} \label{intro_ex2} 
\item "Anna$_i$ told Susan$_k$ to look out the window. She$_{i/k}$ saw that the rain had stopped."
\end{exe}

Who saw that the rain had stopped, Anna or Susan? Could you make an argument for either interpretation? Both possibilities are conceivable, and without additional context, the interpretation is not straightforward.

Most work on coreference resolution focuses on performance in linking each pronoun to a single referent, implicitly assuming that such a referent always exists. In contrast, we use the AmbiCoref dataset and human experiment results from \citet{yuan2023ambicorefevaluatinghumanmodel} to ask how LLMs resolve coreference in comparison with humans, in the presence and in the absence of genuine ambiguity.  We measure (i) how often models choose the same referent that humans prefer in unambiguous sentences and (ii) how often models withhold a choice when the reference is genuinely ambiguous, thereby quantifying the trade-off between performance on unambiguous coreference resolution and the ability to identify unresolvable ambiguity.

\begin{itemize}[leftmargin=*]
\item \textbf{Alignment}: To what extent do LLMs align with human preferences in coreference resolution across both unambiguous and ambiguous contexts?

\item \textbf{``Correct'' accuracy}: How accurately do LLMs resolve coreference in unambiguous sentences, where a single interpretation is strongly preferred by human respondents?

\item \textbf{``Detect'' ambiguity}: Can LLMs reliably distinguish between sentences with sufficient contextual bias to resolve ambiguity and those that remain truly ambiguous for human respondents?

\item \textbf{Balance}: Are LLMs capable of simultaneously achieving high accuracy in unambiguous cases while appropriately detecting ambiguity when present -- a foundational aspect of human-like sentence processing?
\end{itemize}

Our study is the first to our knowledge to compare LLM behavior to 
human judgments specifically in unambiguous \textit{and} ambiguous cases of coreference resolution, and to find the trade-off in disambiguation performance and ambiguity detection in LLMs' base capabilities.

\begin{table*}[!t]
\centering
\small
\begin{tabularx}{\textwidth}{c|X|X|c|c|c}
\toprule
\textbf{Category} &
\textbf{Unambiguous example} &
\textbf{Ambiguous example} &
\textbf{\# Unamb.} &
\textbf{\# Amb.} &
\textbf{Total} \\
\midrule
ECO &
``Matthew told Joshua that he rewarded the client.'' &
``Ruth told the aunt that she baffled the granddaughter.'' &
331 & 332 &  663 \\

ECS &
``William told Joshua that the saleswoman visited him.'' &
``Matthew told Joshua that the client bored him.'' &
131 & 129 &  260 \\

IC &
``Matthew emailed Joshua because he wanted to ask a question.'' &
``The sister-in-law texted Amanda because she is moving abroad soon.'' &
260 & 261 &  521 \\

TOP &
``Matthew wrote Joshua a short poem before he invited him to compose an original verse.'' &
``The sister-in-law sent Amanda a message before she reached the library.'' &
240 &  246 & 486 \\
\hline
\multicolumn{3}{r|}{\textbf{Grand total}} & 962 & 968 &  1930 \\
\cline{4-6} 
\end{tabularx}
\caption{Example sentences and corresponding instance counts for each category in the \textsc{AmbiCoref} dataset.}\vspace{-0.4cm}
\label{tab:combined}
\end{table*}

\section{Related Work}
Ambiguity detection in LLMs remains challenging. \citet{zhang-etal-2024-clamber} show that models like GPT-3.5 and Llama detect ambiguity only slightly above chance, though chain-of-thought prompting improves this marginally. \citet{kim-etal-2024-aligning} attempt to balance ambiguity detection and accuracy in question answering by finetuning models on a dataset labeled based on the model’s own perceived ambiguity.  
Others use re-prompting or interaction with the model to guide its efforts in task disambiguation \citep{niwa2024ambignlgaddressingtaskambiguity}, conversational disambiguation \citep{rahmani2023surveyaskingclarificationquestions, zhang2023clarifynecessaryresolvingambiguity}, question-answering disambiguation \citep{kim-etal-2023-tree,kim-etal-2024-aligning}, entailed meaning disambiguation \citep{liu-etal-2023-afraid},
and questions of lexical, syntactic, and semantic ambiguity \citep{ortegamartín2023linguisticambiguityanalysischatgpt, qamar2024bigclaimslowoutcomes}.
Others investigate models' out-of-the-box ability to handle ambiguous cases by adapting psycholinguistic experiments \citep{cai2024largelanguagemodelsresemble}, following the general example of investigating LLM behavior using psycholinguistics studies \citep{seminck-amsili-2017-computational,ettinger-2020-bert}.

The resolution of ambiguity in sentence processing often relies on semantic cues or on incorporating greater world-knowledge. LLMs' sensitivity to semantic cues is unreliable: while the semantic cues can effectively change the human interpretation of sentences, models have variable responses to this bias, with GPT-based models showing very low sensitivity \citep{lee-etal-2024-multilingual, Scheinberg2025MissingTC}. \citet{lee-etal-2025-relies} find that though the models are able to correctly use world knowledge bias in unambiguous cases, they overextend English syntactic patterns to ambiguous cases in other languages.

On coreference resolution, LLMs have shown strong zero- or few-shot performance, for example, in the Winograd Schema \cite{brown2020languagemodelsfewshotlearners, le2023largelanguagemodelsrobust, gan-etal-2024-assessing}, and typically resolve even ambiguous inputs confidently unless explicitly prompted to express uncertainty \cite{zhang-etal-2024-clamber}. 
Prompting can guide models to resolve pronouns or rewrite text with explicit referents \cite{liu2025bridgingcontextgapsleveraging}, but standard benchmarks like OntoNotes \cite{pradhan-etal-2012-conll} again provided only single-reference annotations and omit ambiguity labels. 
AmbiCoref \cite{yuan2023ambicorefevaluatinghumanmodel} introduced template-generated minimal pairs of unambiguous and ambiguous sentences, showing that humans adjust confidence while coreference models do not. 
A subset of this dataset includes human-annotated responses, which we use for direct comparison with the LLMs’ performance, bypassing specialized coreference resolution models as benchmarks since our focus is on human-like processing. This approach allows us to investigate both the ability of LLMs to determine coreferents and the pattern of how ambiguity impacts their output.

\section{Experimental Setup}

\subsection{Dataset}

We use the AmbiCoref dataset \citep{yuan2023ambicorefevaluatinghumanmodel} to investigate how coreference resolution models’ behavior and human annotations differ in unambiguous and ambiguous cases.
Each sentence in the dataset has a first clause that mentions two persons, and a second clause that contains a pronoun referring to one of them. A question paired with each sentence can serve to resolve the ambiguity (e.g. ``who saw that the rain had stopped'' queries the referent in Example \ref{intro_ex2}). AmbiCoref contains 968 ambiguous and 962 unambiguous sentences of different categories (see Table  \ref{tab:combined}).

\begin{table*}[!ht]
\centering
\footnotesize
\ttfamily
\begin{tabularx}{\textwidth}{X}
\toprule
\textbf{\textsc{Reflect} Prompt} \\
\midrule

You will be presented a sentence. 
Each sentence contains at least two nouns, and a nominative third-person singular pronoun (he/she).
If three nouns appear in the sentence, please only consider the first two nouns as possible candidates.

Following each sentence, there will be a question asking you which noun the specified pronoun refers to, or if you find the pronoun ambiguous.

For example, \textbf{Chloe told Emma that she was sad}.\textit{ Who was sad?}

\makebox[1em]{--} \textit{Definitely Chloe} - I feel confident that "she" refers to "Chloe";

\makebox[1em]{--} \textit{Probably Chloe} - I think "she" refers to "Chloe" as opposed to "Emma" but I feel unsure;

\makebox[1em]{--} \textit{Mostly ambiguous} - I find it completely ambiguous whether "she" refers to "Chloe" or "Emma";

\makebox[1em]{--} \textit{Probably Emma} - I think "she" refers to "Emma" as opposed to "Chloe" but I feel unsure;

\makebox[1em]{--} \textit{Definitely Emma} - I feel confident that "she" refers to Emma.

Different people may have different judgments and tolerance for ambiguity, so please feel free to use your intuitive judgments.

<sentence>

<question>
\\\bottomrule
\end{tabularx}
\caption{\textsc{Reflect} instruction prompt.} 
\label{tab:sys_role}
\end{table*}
{Each sentence falls into one of four categories, distinguished by semantic properties associated with the verb.}
The {\em unambiguous} subset restricts the verb or phrase to one plausible meaning
\footnote{These restrictions cause a bias toward one of the referents over the other, however, there is still a degree of ambiguity in these `unambiguous' sentences. Additional context could arguably change the preferred reading of the sentences, and bias the coreference resolution to the other referent. The dataset items are only at sentence-level, so here there is no greater context to disrupt the bias in the sentence itself.}, and an {\em ambiguous} subset, where both interpretations are plausible. Please see \citet{yuan2023ambicorefevaluatinghumanmodel} for further details. The categories are:

\noindent\textbf{Experiencer Constraint for Objects (ECO)}: The object of the sentence is constrained by the verb to be interpreted as the experiencer of that verb. 

\noindent\textbf{Experiencer Constraint for Subjects (ECS)}: The subject of the sentence is constrained by the verb as its experiencer. 

\noindent\textbf{Implicit Causality (IC)}: The verb implies a causality that constrains the coreferent. 

\noindent\textbf{Transfer of Possession (TOP)}: The verb determines the coreferent according to the logic of the source-goal transfer in the sentence.

\begin{figure*}[t!]
\centering
\begin{subfigure}[b]{0.35\linewidth}
\centering
\includegraphics[width=\linewidth]{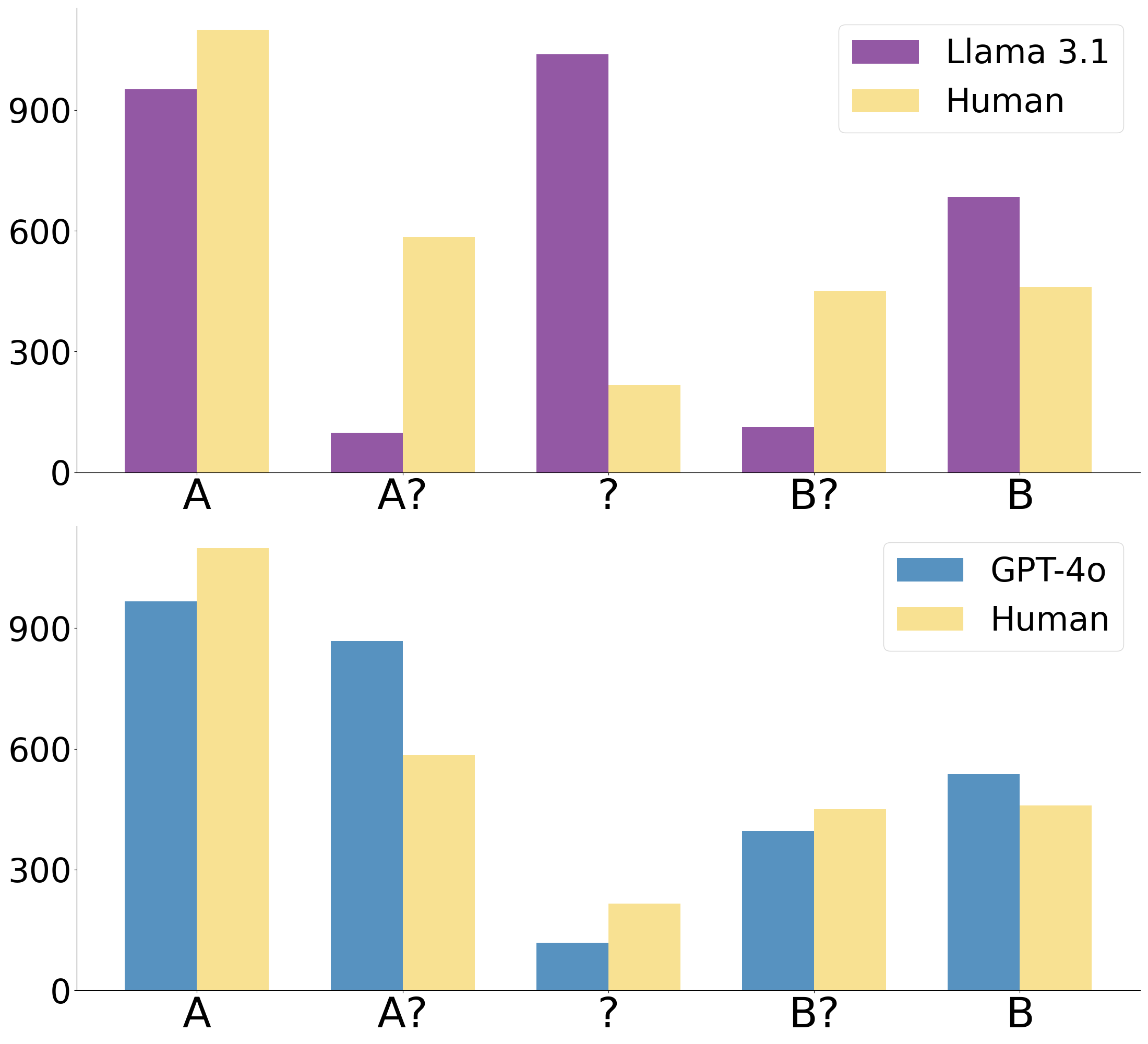}
\caption{Unambiguous}
\label{fig:stacked_unambiguous}
\end{subfigure}
\hspace{0.1\linewidth}
\begin{subfigure}[b]{0.35\linewidth}
\centering
\includegraphics[width=\linewidth]{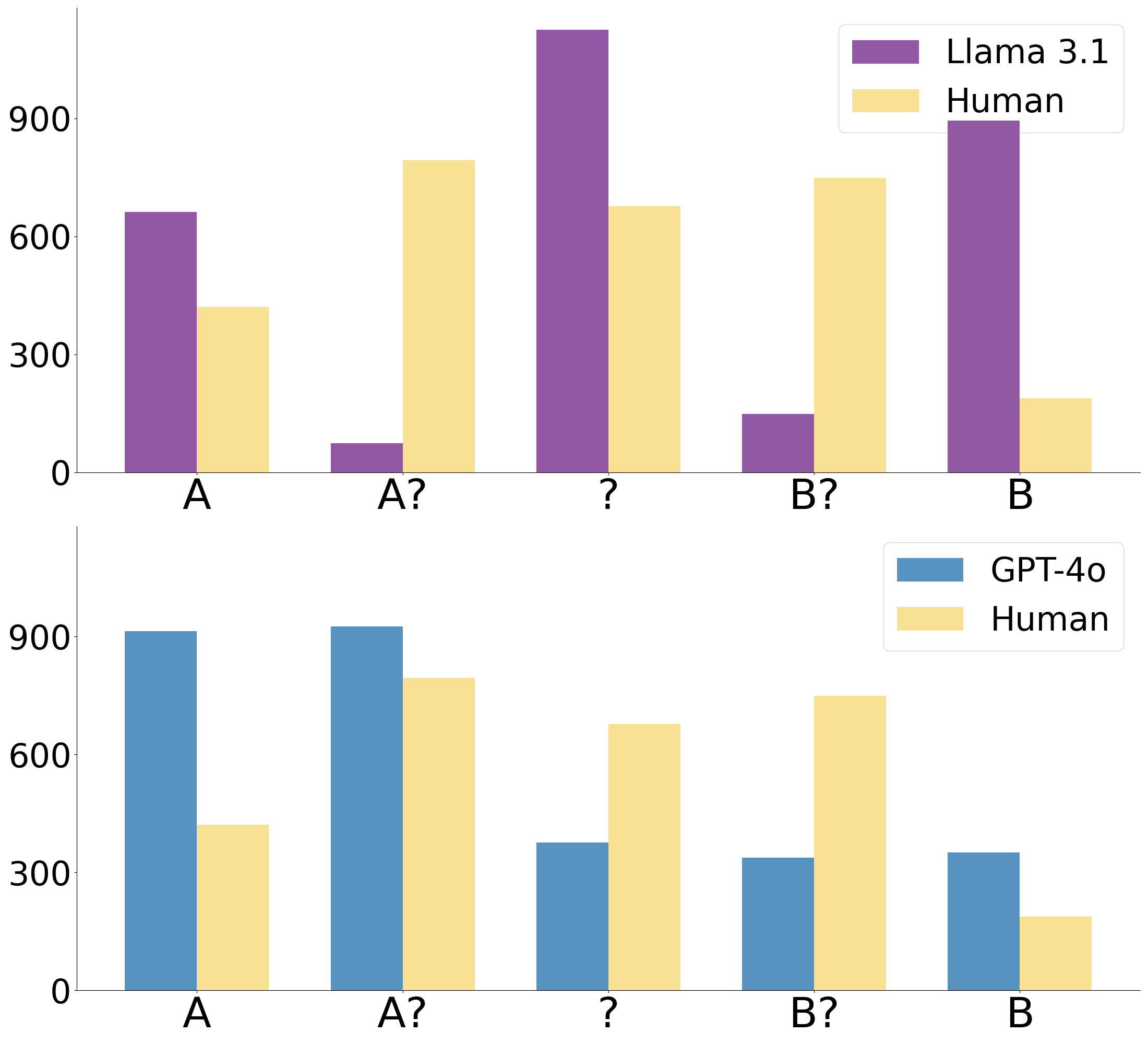}
\caption{Ambiguous}
\label{fig:stacked_ambiguous}
\end{subfigure}
\caption{
    \textsc{Reflect}: Model results compared to human answer patterns across unambiguous and ambiguous sets. Human responses exhibit a U-shaped distribution for unambiguous sentences, with a decreased preference for `ambiguous' and `probably' answers (?, A?, B?). This pattern reverses for ambiguous sentences, forming an inverted U-shaped distribution due to an increased preference for the `ambiguous' (?) responses. The models do not strongly track this pattern: Llama 3.1 fails to adjust in the ambiguous set, while GPT-4o slightly increases `?' responses.
}\vspace{-0.4cm}
\label{fig:stacked}
\end{figure*}

\subsection{Models}
We evaluate two LLMs, \textbf{GPT-4o} (\texttt{gpt-4o-2024-08-06}) \citep{openai2024gpt4technicalreport} and \textbf{Llama 3.1 70B} (\texttt{Llama3.1-70b-Instruct}) \citep{dubey2024llama3herdmodels}. We chose GPT-4o to reflect the state-of-the-art in LLM performance and we chose Llama 3.1 as a comparable open-source model. 
We tested each model over three runs on all 1,930 sentences, collecting a total of 5,790 responses per model.

\section{\textsc{Reflect}ing Human Patterns in Coreference Resolution}
Initially, we adopt the text originally presented to human subjects to test how the language models reflect human preferences; this \textsc{Reflect} prompt is shown in Table \ref{tab:sys_role}. We observed an order bias in preliminary results similar to that reported in \citet{pezeshkpour2023largelanguagemodelssensitivity} so determined not to provide a list of possible referents, in line with \citet{ide2024makellmsgrammaticalknowledge}. 

In this section we compare model results with the human judgments. Model results are compared with human annotations by mapping both to the schema
[\textbf{A},
\textbf{A?},
\textbf{?},
\textbf{B?},
\textbf{B}], 
where \textbf{A} is the label for the first person in the sentence and \textbf{B} is the second.  In the schema, \textbf{?} indicates a degree of ambiguity: while \textbf{A} corresponds to ``Definitely Chloe'', \textbf{A?} corresponds to ``Probably Chloe'' and \textbf{?} indicates ``Mostly ambiguous''.

\begin{table*}[!t]
\centering
\small
\setlength{\tabcolsep}{8pt}  
\begin{tabular}{lcccc|c|cccc|c|c}
\toprule
\multirow{2}{*}{} & \multicolumn{4}{c|}{\bf Unambiguous} & \multirow{2}{*}{Avg} & \multicolumn{4}{c|}{\bf Ambiguous} & \multirow{2}{*}{Avg} & \multirow{2}{*}{$\Delta$} \\
\cmidrule{2-5} \cmidrule{7-10}
& ECO & ECS & IC & TOP & & ECO & ECS & IC & TOP & & \\
\midrule
Human & 24.12 & 29.84 & 21.20 & 11.26 & 21.61 & 11.73 & 11.38 & 8.40 & 6.01 & 9.38 & 12.23 \\
Llama 3.1 & 25.98 & 19.85 & 28.46 & 28.33 & 25.66 & 20.78 & 21.71 & 22.99 & 28.86 & 23.59 & 2.07 \\
GPT-4o & 51.06 & 42.75 & 44.62 & 35.42 & 43.46 & 47.89 & 34.11 & 41.38 & 38.21 & 40.40 & 3.06 \\
\bottomrule
\end{tabular}
\caption{Answer consistency percentages across unambiguous and ambiguous conditions for different models. $\Delta$ (difference between unambiguous and ambiguous averages) highlights the responsiveness to ambiguity. Human average answer consistency shifts about four times as much as the models do in response to ambiguity.}
\label{tab:unanimous_answer}
\end{table*}

\begin{table}[!t]
\centering
\setlength{\tabcolsep}{14pt}
\small
\begin{tabular}{l c c }
\toprule
  & \multicolumn{2}{c}{\textbf{Accuracy}}\\
& strict & near \\
\midrule
Human      & 50.94 & 76.77\\
Llama 3.1     & 38.39 & 43.07\\
GPT-4o      & 43.38 & 79.52\\
\bottomrule
\end{tabular}
\caption{Accuracy (strict and near) on unambiguous instances using \textsc{Reflect} prompt} \vspace{-0.4cm}
\label{tab:sub-correct}
\end{table}

\paragraph{\em Models Inconsistently Reflect Unambiguous Answer Patterns} {Figure \ref{fig:stacked_unambiguous}} presents the results of coreference resolution preferences of models and humans side by side for the unambiguous sentences. Our results indicate that Llama 3.1 shows poor alignment with human results, particularly in having a much greater proportion of \textbf{?} responses (``mostly ambiguous'') 
and in dispreferring \textbf{A?} and \textbf{B?} (the ``probably'' answers).
On the other hand, we observe that GPT-4o's preferences align much better with human decisions in the unambiguous set. 
See Appendix \ref{app:model_patterns} for further discussion of model patterns broken down by sentence category and the by-category chi-square test for statistical significance for these results.

\paragraph{\em Models Show Minimal Behavior Shift between Unambiguous and Ambiguous Sentences} Human annotators' preferences show markedly different patterns between unambiguous and ambiguous sentences, indicating that humans are sensitive to the \textit{lack} of semantic constraints in the ambiguous case. Do LLMs show the same responsiveness?

On the contrary, Figure \ref{fig:stacked_ambiguous} indicates a substantially smaller shift in coreference resolution preference as compared with the shift observed in human subjects -- thus we find a further divergence between LLM and human disambiguation behaviors. In Appendix \ref{app:fullanswers} we further analyze this divergence by the categories.

\paragraph{\em Answer Consistency Shifts are Low in Models, High in Human Responses} \label{sec:answer_consistency}
Apart from alignment with or divergence from human preferences, we examine each model's response consistency on the unambiguous or ambiguous stimuli, that is, how often each model's three responses to a sentence are the same or different. Intuitively, we expect per-sentence answer consistency to be higher across runs in the unambiguous subset, and to show more variation in the ambiguous subset.

We investigate by model and between human annotators and present the results in Table \ref{tab:unanimous_answer}. The data confirm our predictions for the human results: 
the human annotators choose the same answer 21.61\% of the time for unambiguous sentences and only 9.38\%
of the time for ambiguous sentences, showing 1) a relatively high consistency for five classes in the unambiguous set, and 2) a reduction in consistency of over 12 percentage points between unambiguous and ambiguous sentences.

Indeed, we expect that removing bias towards one of the referents should increase the variability in responses. 

On the other hand, models diverge in both points: 1) they are more stable in their judgments: Llama 3.1's consistency is 25.66\% in the unambiguous and 23.59\% in the ambiguous sentences, and GPT-4o shows a much higher consistency, averaging 43.46\% in the unambiguous subset with a drop to 40.40\% in the ambiguous subset, and 2) models shift only marginally in consistency between unambiguous and ambiguous data, whereas human annotators' consistency is markedly lower for ambiguous data.

\paragraph{\em Accuracy on Unambiguous Sentences with \textsc{Reflect} prompt}
The semantic constraints in AmbiCoref are designed to \textit{influence} interpretation of which entities are co-referents. While these `unambiguous' sentences are biased towards a particular resolution rather than absolute, we define accuracy as alignment with these semantically biased interpretations, treating them as ground truth labels. 

Two answer schemas are used in our analysis: 5-answer (strict) and 3-answer (near). 
While 5-answer measures strict accuracy, the
3-answer schema collapses person mentions even if the labeler is uncertain, resulting in three possible answers: \textbf{[(A, A?)}/\textbf{?}/\textbf{(B?, B)]}, so it measures near-correctness.

Under the strict metric, we observe poor overall performance (see Table~\ref{tab:sub-correct}), with GPT-4o scoring 43.38\% and Llama 3.1 following at 38.39\%, suggesting that current models struggle to confidently identify the intended referent even when semantic cues strongly favor one interpretation. 
The human annotations achieve 50.94\% in strict correctness, and 76.77\% in near-correctness. 
Widening our correctness criteria to near-correctness helps the models as well: 43.07\% for Llama 3.1 and 79.52\% for GPT-4o.

\paragraph{\em Model Explanations Can Indicate Challenging Items}
Our prompting approach for the models emulates the human annotation instructions for strong comparability. 
This approach also led to large variability in model outputs: most notably, models sometimes volunteered explanations for their choices (which we did not explicitly request). In both Llama 3.1 and GPT-4o, we find that responses with word counts greater than 20 indicate more than a simple answer statement.
In responses not longer than 20 words, model responses simply \textbf{report} their choice (example response (19 words): ``\textit{Mostly ambiguous. It's unclear whether `she' refers to `Melissa' or `Jennifer', as both interpretations are possible in this sentence.}''), but longer responses \textbf{explain} their choice, citing sentence structure, hypothetical reasoning, and discussions of ambiguity.

\begin{table}[!t]
\centering
\small
\setlength{\tabcolsep}{3pt}
\begin{tabular}{l | ccc | cc}
\toprule
\textbf{Model} & 
\multicolumn{3}{c|}{\textbf{\% Explain}} & 
\multicolumn{2}{c}{\textbf{Accuracy (Unamb)}} \\
& \textbf{Overall} & \textbf{Unamb} & \textbf{Amb} & \textbf{No-Explain} & \textbf{Explain} \\
\midrule
Llama 3.1   & 67.53 & 67.08 & 67.98 & 39.70 & 37.80 \\
GPT-4o      & 12.75 & 10.91 & 14.57 & 45.10 & 29.50 \\
\bottomrule
\end{tabular}
\caption{\textbf{\textit{Explain}}: Unprompted explanation rates and accuracy across unambiguous and ambiguous examples. The responses with an explanation in Llama 3.1 have no significant relationship to accuracy, while in GPT-4o they are correlated with lower accuracy. } 
\label{tab:explain_correct}
\end{table}

\begin{table}[!t]
\centering
\small
\setlength{\tabcolsep}{14pt}
\begin{tabular}{lcc}
\toprule
& \multicolumn{2}{c}{\textbf{Accuracy (\%)}} \\ 
& \textbf{Female} & \textbf{Male} \\
\midrule
Human      & 49.14 & 52.86 \\
Llama 3.1   & 29.34 & 47.90 \\
GPT-4o      & 34.89 & 52.31 \\
\bottomrule
\end{tabular}
\caption{\textbf{Gender bias}: Accuracy on unambiguous sentences containing female- vs.\ male-gendered pronouns and names. Accuracy for human answers is mostly consistent, and models show worse accuracy in sentences with female pronouns and names.} 
\label{tab:gender_accuracy}\vspace{-0.5cm}
\end{table}

The results are presented in Table~\ref{tab:explain_correct}. About 68\% of all Llama 3.1 responses (unambiguous and ambiguous) contain some amount of explanation offered for the answer, evenly split between unambiguous and ambiguous sentences, with no significant differences between sentence categories. 
In contrast, GPT-4o over-answers less frequently (only 13\%), with a higher tendency to do so in ambiguous cases than in unambiguous cases, and shows meaningful differences by categories (for a per-category investigation, see Appendix \ref{app:explain}).

Focusing on the unambiguous set, for Llama 3.1, \textit{explain} percentages stay consistent over the model's performance:
the model does not show a significant difference in accuracy in \textit{explain} responses (37.8\%) versus \textit{no-explain} responses (39.7\%). However, GPT-4o's responses show a different pattern: explanations are an indication of poor performance. 
Within the \textit{explain} responses, it has an accuracy of 29.5\%, versus the 45.1\% accuracy in \textit{no-explain} responses. GPT-4o's performance here shows an inverse correlation between the presence of an explanation and the model's performance. 

\paragraph{\em Gender Bias Still Hinders Performance}
As noted by \citet{davis-van-schijndel-2020-discourse}, gender biases in neural models can be made visible in coreference resolution. While they looked at gender combined with model surprisal at the coreferring pronoun, we can see the effects in overall performance. The dataset is almost evenly split between sentences that use only female pronouns and names (51\%) versus sentences with only male pronouns and names (49\%). They do not mix these in the same sentence.

While humans had very similar accuracy in both gendered subsets (49.14\% for female pronouns and 52.86\% for male pronouns), models differed significantly (Table \ref{tab:gender_accuracy}). 
Llama 3.1 attained an accuracy of 29.34\% (female) and 47.9\% (male), while GPT-4o showed a similar difference with 34.89\% (female) and 52.31\% (male). Male pronouns and culturally masculine names correlate with better model accuracy. With gender bias a known issue in LLMs, we are disappointed but not surprised by this finding.

\begin{table}[!t]
\centering
\begin{minipage}{0.48\textwidth}
\footnotesize
\ttfamily
\setlength{\tabcolsep}{2pt}   
\begingroup
\setlength{\extrarowheight}{.6em}
\begin{tabularx}{\textwidth}{@{}lX@{}}
\toprule
\textbf{Template} & \textbf{Prompt text} \\
\midrule
\textsc{Basic} &
<sentence>\newline
<question> \\

CoT & 
Show your reasoning step-by-step, then provide your answer as "A:".\newline
<sentence>\newline
<question> \\

WSC & 
<sentence>\newline
Who is <pronoun> referring to? \\

WSC-1sh &
Sentence: Hannah told Jasmine that she impersonated Natalie.\newline
Question: Who is "she" referring to?\newline
Answer: The pronoun "she" is referring to Hannah.\newline
Sentence: <sentence>\newline
Question: Who is <pronoun> referring to?\newline
Answer: \\
\bottomrule
\end{tabularx}
\endgroup
\caption{Prompts for Coreference Resolution.} 
\label{tab:correct_full_prompts}\vspace{-0.4cm}
\end{minipage}
\end{table}

\setlength{\tabcolsep}{18pt}
\begin{table*}[t]
\centering
\small
\begin{tabular}{l l r r r r r}
\toprule
   & \textbf{Prompt}   & \textbf{All} & \textbf{ECO} & \textbf{ECS} & \textbf{IC} & \textbf{TOP} \\

\midrule
Human     & \textsc{Reflect} (near)  & 76.77 & 79.75 & 82.81 & 78.62 & 67.29 \\
\midrule
Llama 3.1 & \textsc{Reflect} (near)    & 43.07 & 46.32 & 23.41 & 56.03 & 35.28 \\
Llama 3.1 & \textsc{Basic}  & \textbf{90.33} & \textbf{94.76} & \textbf{96.44} & \textbf{98.85} & 71.67 \\
Llama 3.1 & CoT   & 86.38 & 84.29 & 88.8 & 96.67 & \textbf{76.81} \\
Llama 3.1 & WSC     & 39.50 & 42.40 & 73.54 & 26.28 & 31.25 \\
Llama 3.1 & WSC-1sh & 69.82 & 80.06 & 39.19 & 66.79 & 75.69 \\
\midrule
GPT-4o   & \textsc{Reflect} (near)     & 79.52 & 91.14 & 64.63 & 88.08 & 62.36 \\
GPT-4o   & \textsc{Basic}   & 87.70 & 89.83 & 88.55 & 96.54 & \textbf{74.72} \\
GPT-4o   & CoT    & \textbf{89.99} & \textbf{95.17} & \textbf{93.38} & \textbf{97.44} & 72.92 \\
GPT-4o   & WSC      & 39.43 & 28.80 & 38.93 & 47.82 & 45.28 \\
GPT-4o   & WSC-1sh  & 45.43 & 62.44 & 38.68 & 30.90 & 41.39 \\
\bottomrule
\end{tabular}
\caption{\textsc{Correct} results: Accuracy on unambiguous instances for coreference resolution prompt types. The highest scores are found in the \textsc{Basic} and CoT settings.} \vspace{-0.4cm}
\label{tab:baselines}
\end{table*}

\section{\textsc{Correct} Accuracy on Coreference Resolution}

When released from the constraint of reflecting human linguistic behavior, how do models fare on accuracy in the unambiguous subset?

\paragraph{\em Experiments}
In addition to our human-comparison \textsc{Reflect} prompt, we experiment with several other prompts (Table \ref{tab:correct_full_prompts}). We provide an instruction-less prompt (\textbf{\textsc{Basic}}) in line with previous research that found enhanced response quality in unrestricted output \citep{chiang2023closerlookautomaticevaluation,gan-etal-2024-assessing}. For comparison with standard coreference resolution models, we also experiment with a Winograd schema \citep{levesque-wino-2012} style question directly (\textbf{WSC}) and in a one-shot setting (\textbf{WSC-1sh}). While our task is not structured like a traditional coreference resolution task, we can follow the example of recent work \citep{gan-etal-2024-assessing} that has explored how to adapt the use of LLMs to this more-functional version of the task. 
The best results in that work used Chain-of-Thought style prompting (\textbf{CoT}) and so we adapt that to this task as well.

\paragraph{\em Results}
Table \ref{tab:baselines} shows accuracy results on the unambiguous items from the dataset. These results point to the issue posed by too much instruction, as we compare the performance of the \textsc{Reflect} prompt to the \textsc{Basic} prompt. Instructions allow for more structured, easily parsable answers from the model, but they can introduce bias that can be hard to account for.

The WSC-style prompts are a poor framing for this task overall, though the one-shot version (WSC-1sh) is able to gain an overall improvement in both models. It helps most for Llama 3.1, boosting overall scores from 39.5\% to 69.82\%. In the WSC setting, Llama 3.1 has many instances where it misinterprets the question as one of entity disambiguation, answering with references to famous people or figures from literature.

Overall, CoT and \textsc{Basic} prompts are the best performing prompts. CoT-style prompts lead to a large number of tokens in the output (GPT-4o has an average response word count of 135 in this setting, and Llama has an average of 83). When considering efficiency, the \textsc{Basic} prompt responses are clearly preferred, with an average word count of 14, and we build on this as our best method.

\section{Can We Balance Coreference Resolution Performance and Ambiguity Detection?}
\label{sec:model_behavior}
Introducing genuine ambiguity turns coreference from a single-metric task to a dual challenge: a model should choose a referent when context singles one out, yet abstain when context leaves the choice open. Achieving both goals simultaneously may not be possible, because, as we will see, gains in one often cost the other -- a trade-off similar to those of invariance vs. adaption \citep{lucy-etal-2024-system-behaviors} and  fairness vs. accuracy \citep{dutta-et-al-tradeoff}.

We demonstrate another trade-off with the \textsc{Correct}-\textsc{Detect} experiments presented below.

\begin{table}[!t]
\centering
\begin{minipage}{0.48\textwidth}
\footnotesize
\ttfamily
\setlength{\tabcolsep}{3pt}
\begingroup
\begin{tabularx}{\textwidth}{@{}lX@{}}
\toprule
\textbf{Template} & \textbf{Prompt text} \\
\midrule
Ambi-Ask &
Answer the question below. If you find the question ambiguous, answer “Ambiguous” instead.\newline
Sentence: <sentence>\newline
Question: <question>\\
\midrule
Ambi-Stop &
Sentence: <sentence>\newline
Question: <question>\newline
Stop to consider: is there enough context to answer confidently? If yes, then answer. If no, then answer "Ambiguous." \\
\midrule
Ambi-Wait &
Sentence: <sentence>\newline
Question: <question>\newline
Wait, can the ambiguity here be resolved given the context you have? If yes, answer only with the person, and if no, answer "ambiguous." \\
\midrule
Ambi-Doubt &
Sentence: <sentence>\newline
Question: <question>\newline
Do you have any doubt as to the answer to this question? If yes, answer "ambiguous," or else reply only with the person who is the answer. \\
\midrule
Ambi-CoT &
Sentence: <sentence>\newline
Task: Identify "<question>", or say Ambiguous if unclear.\newline
Think step-by-step:\newline
First, consider the people in the sentence. Then determine if only one could logically be the answer within this context. If there are multiple equally possible candidates, then the sentence is ambiguous. Finally, give the answer: state the correct person, or state `Ambiguous.'\\
\bottomrule
\end{tabularx}
\endgroup
\caption{Full text of the \textsc{Correct}-\textsc{Detect} prompts.} 
\label{tab:detect-prompts}
\end{minipage}
\end{table}

\paragraph{\em Experiments}
We report two complementary metrics, \textbf{Correct-Unamb} and \textbf{Detect-Ambig}. Correct-Unamb indicates accuracy on the unambiguous set, and Detect-Ambig measures detection of ambiguity in ambiguous sentences (\textbf{?}, \textbf{A?} and \textbf{B?} all count as detections). High scores on both metrics would show that a model both solves clear cases and recognizes genuine ambiguity. 

Building on the \textsc{Basic} prompt, we create five variants which highlight the possibility of ambiguity in different ways, in order to test whether a model can be both accurate and sensitive to ambiguity: \textbf{Ambi-Ask} directly asks about ambiguity, \textbf{Ambi-Stop} and \textbf{Ambi-Wait} use the word `stop' or `wait,' \textbf{Ambi-Doubt} asks about doubt, and \textbf{Ambi-CoT} prompts the LLM to think ``step-by-step''. See the full prompt texts in Table \ref{tab:detect-prompts}.

\begin{figure}[!t]
\centering
\includegraphics[width=\linewidth]{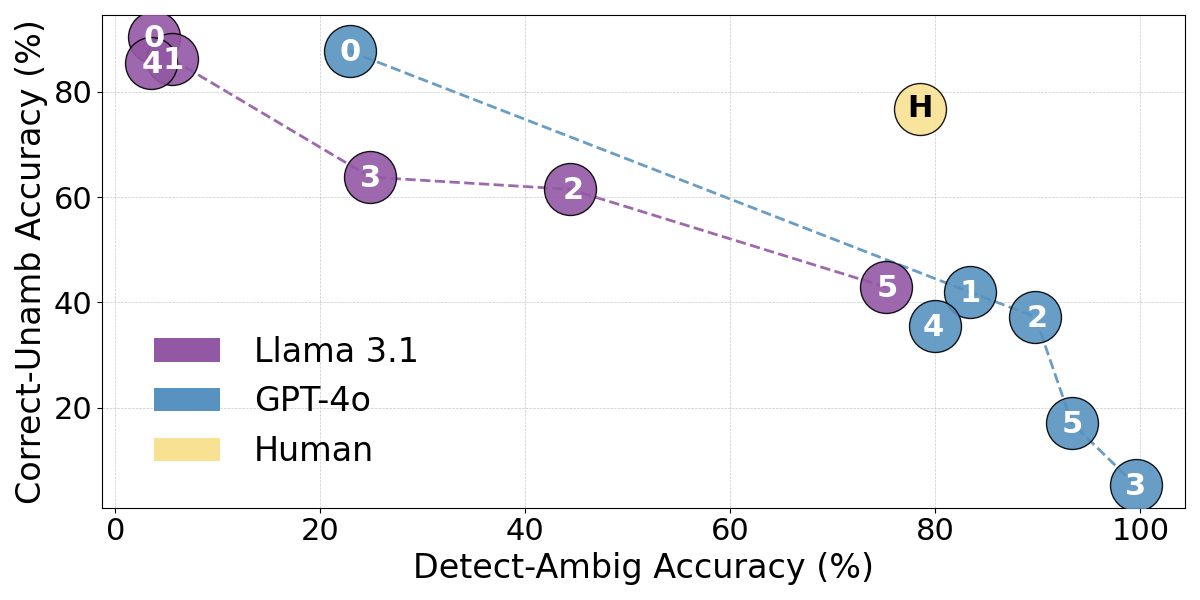}

\caption{\textbf{Detect-Ambig plotted against Correct-Unamb.} 0: \textsc{Basic}, 1: Ambi-Ask, 2: Ambi-Stop, 3: Ambi-Wait, 4: Ambi-Doubt, 5: Ambi-CoT. H shows the human results for near-correctness.} 
\label{fig:pareto}
\end{figure}

\paragraph{\em Results}
The results are presented in Table \ref{tab:ambiack} and in Figure \ref{fig:pareto}, which shows the Pareto front of all relevant experiment settings indicating a trade-off. 

The human annotators balance the task well, identifying ambiguity in 78.47\% of the ambiguous sentences, while correctly answering 76.77\% of the unambiguous sentences.

The \textsc{Basic} prompt allows for the models to achieve high accuracy in answering the unambiguous sentences (90.33\% for Llama 3.1 and 87.7\% for GPT-4o), however the detection rates for the ambiguous sentences are the lowest under this setting (3.72\% for Llama 3.1 and 22.86\% for GPT-4o).

Llama 3.1 is unable to find a good balance: four out of five of the \textbf{Ambi-}prompts do worse than chance on detection, even though accuracy stays high. GPT-4o is excellent at ambiguity identification, reaching an astonishing 99.55\% with the \textbf{Ambi-Wait} prompt, but sacrificing accuracy on unambiguous cases: the same \textbf{Ambi-Wait} prompt has only 5.23\% accuracy.

Notably, the same prompting strategy can elicit very different results in different models: GPT-4o achieves the best balance with \textbf{Ambi-Ask} (with 83.37\% detection and 41.93\% accuracy), while \textbf{Ambi-Ask} elicits Llama 3.1's worst detection score (5.54\%) but its \textit{best} accuracy on unambiguous items (86.17\%).

\begin{table}[!t]
\centering
\setlength{\tabcolsep}{6pt}
\small
\begin{tabular}{l l r r}
\hline
   & \textbf{Prompt} & \textbf{Detect} & \textbf{Correct} \\
\toprule
Human       & \textsc{Reflect} (near)        & 78.47 & 76.77 \\
\midrule
Llama 3.1   & \textsc{Basic}          & 3.72 & 90.33\\
Llama 3.1   & Ambi-Ask       & 5.54 & 86.17\\
Llama 3.1   & Ambi-Stop      & 44.35 & 61.47\\
Llama 3.1   & Ambi-Wait      & 24.86 & 63.79\\
Llama 3.1   & Ambi-Doubt     & 3.51 & 85.41\\
Llama 3.1   & Ambi-CoT       & 75.17 & 42.90\\
\midrule
GPT-4o      & \textsc{Basic}          & 22.86 & 87.70\\
GPT-4o      & Ambi-Ask       & 83.37 & 41.93\\
GPT-4o      & Ambi-Stop      & 89.74 & 37.28\\
GPT-4o      & Ambi-Wait      & 99.55 & 5.23\\
GPT-4o      & Ambi-Doubt     & 79.94 & 35.45\\
GPT-4o      & Ambi-CoT       & 93.31 & 17.08\\
\bottomrule
\end{tabular}
\caption{Correct answers in the Unambiguous case (\textbf{Correct-Unamb}), and detecting ambiguity in the Ambiguous case (\textbf{Detect-Ambig}).} 
\label{tab:ambiack}
\end{table}

\paragraph{\em Trade-off in Ambiguity Detection vs. Coreference Resolution Performance}
Across all prompts we observe a persistent tension: high performance in \textbf{Correct-Unamb} tends to be accompanied by lower performance in \textbf{Detect-Ambig}, and vice versa. No model appears in the Pareto plot's top-right quadrant, which represents high scores on both metrics. Prompting the model to think about ``ambiguity,'' ``doubt,'' or ``confidence'' reliably boosts \textit{Detect-Ambig} but reduces accuracy on unambiguous items. Prompts that simply demand a coreference resolution show high accuracy but leave ambiguity undetected.

{We also investigate whether a model designed for reasoning tasks can do better on the initial AmbiCoref dataset. We report results on OpenAI's o3  reasoning model \cite{openai-o3-2025} in Table \ref{tab:reasoning_results}. These results fall into the same pattern as the previous models: higher ambiguity detection scores (73.17\% and 84.47\%) are paired with lower accuracy (41.55\% and 36.04\%). These results reflect the trade-off pattern seen in the CoT-based prompt in Table \ref{tab:ambiack}, where reasoning improves ambiguity detection at the cost of accuracy.}

\begin{table}[!t]
\centering
\small
\setlength{\tabcolsep}{6pt}
\begin{tabular}{l l r r}
\toprule
   \textbf{Prompt} & \textbf{Correct} & \textbf{Detect} \\
\midrule
Ambi-Ask       & 41.55  & 73.17 \\
Ambi-Stop      & 36.04 & 84.47 \\
\bottomrule
\end{tabular}
\caption{\textbf{Reasoning Model (GPT o3) Results: }Correct answers in the Unambiguous case (\textbf{Correct}), and detecting ambiguity in the Ambiguous case (\textbf{Detect}).} 
\label{tab:reasoning_results}\vspace{-0.4cm}
\end{table}

While this trade-off might be masked by combining models or prompting the model in multiple rounds,
future work should look to improving models themselves to approach or exceed the human benchmark in that upper-right region, providing decisive answers when context suffices and principled abstention when it does not.

\section{Additional Experiments: Ambiguity in Entailment}
AmbiCoref is unique in its combination of features, {comprising both ambiguous and unambiguous items and calling simultaneously for ambiguity detection and accurate resolution}. Because of the lack of other datasets combining these attributes, it is unclear how the tradeoff uncovered here may generalize to other task types, especially more complicated ones. 

A second dataset that does allow for differential handling of unambiguous and ambiguous instances is AmbiEnt from \citet{liu-etal-2023-afraid}, an entailment dataset with ambiguous instances and paired disambiguations. Ambiguity in these premise-hypothesis pairs is indicated by the number of labels: multiple labels represent multiple potential readings of the pair. The pairs are annotated with paraphrases that allow for the selection of one label as correct over the others, disambiguating the entailment label.
Using this dataset, we provide an investigation of the dual-goal performance on a new task, with a subset of our prompt types.

We ran GPT-4o and Llama 3.1 70B on AmbiEnt. We use a subset of the data annotated by linguists for ambiguity type, which includes coreference resolution. After removing items missing labels from the subset annotated with ambiguity types, we have 62 pairs (124 total items) of unambiguous and ambiguous instances.

As reported in Table \ref{tab:ambient}, when asked whether the premise was ambiguous, the models generally do not acknowledge ambiguity at all (scoring between 0\% - 21\% on Detect). Instead they answer as if it were a standard entailment task with no potential ambiguity. This highlights the difficulty in revealing the Correct-Detect tradeoff: the models already do not perform well at identifying ambiguity, so it is difficult to design contexts where optimizing for ambiguity detection can be seen decreasing accuracy.
Expanding the domain where we can test this effect remains part of our future work.

\section{Discussion}

{Our correct-detect trade-off findings have a strong resonance with \citet{kalai2025languagemodelshallucinate}'s analysis that problematic training incentives contribute to factual hallucinations: current training and evaluation procedures reward confident guessing over acknowledging uncertainty, since binary scoring gives no credit for appropriate ``I don't know'' responses. We hypothesize that the incentives underlie LLMs’ inability to combine ambiguity detection with accurate resolution: models are incentivized to guess confidently even when faced with ambiguity. However, while \citet{kalai2025languagemodelshallucinate} show that factual hallucination cannot be entirely avoided, we suspect that linguistic ambiguity detection may be more tractable: if the incentive problem is overcome, we predict that models will quickly cross the current Pareto front in Figure \ref{fig:pareto} and learn to handle linguistic ambiguity in a more human-like way.}

\begin{table}[!t]
\centering
\setlength{\tabcolsep}{6pt}
\small
\begin{tabular}{l l r r}
\toprule
   & \textbf{Prompt} & \textbf{Correct} & \textbf{Detect} \\
\midrule
Llama 3.1   & Ambi-Ask       & 67.74  & 0 \\
Llama 3.1   & Ambi-Stop      & 70.97 & 1.61\\
\midrule
GPT-4o      & Ambi-Ask       & 74.19 & 0 \\
GPT-4o      & Ambi-Stop      & 69.35 & 20.97 \\
\bottomrule
\end{tabular}
\caption{\textbf{AmbiEnt Results: }Correct answers in the Unambiguous case (\textbf{Correct}), and detecting ambiguity in the Ambiguous case (\textbf{Detect}).} \vspace{-0.2cm}
\label{tab:ambient}
\end{table}

\section{Conclusion}

The goal is contextually responsive, decisive, and ambiguity-aware coreference resolution.
While models have demonstrated the ability to achieve high scores on either coreference resolution performance or ambiguity detection depending on the prompt style, these goals compete when combined. 
Models do not strongly reflect human answer patterns in coreference resolution, especially in ambiguous cases. GPT-4o did much better on this measure than Llama 3.1, but neither strongly shift answer patterns in response to ambiguity as humans do.
A sampling of methods shows that it is not currently possible to elicit high performance on both measures simultaneously in models. 

\clearpage
\newpage

\section*{Limitations}
While we chose comparable models, as always the addition of more models would strengthen these results. 
This work is limited to only English language data. Ambiguity in different languages has different considerations and would need to be investigated separately. Also, the types of sentences in the dataset we use are purposefully limited in number of mentions and context length, both of which are directions for potential future work.

\section*{Acknowledgments}
We are grateful to the anonymous reviewers for their constructive feedback which helped improve this paper. We would like to thank the PortNLP lab for their support and  E. Devin Vander Meulen II for designing Figure 1.


\begin{thebibliography}{39}
\providecommand{\natexlab}[1]{#1}

\bibitem[{Bousquet et~al.(2020)Bousquet, Swaab, and Long}]{bousquet2019useofcontext}
Kathryn Bousquet, Tamara~Y Swaab, and Debra~L Long. 2020.
\newblock \href {https://doi.org/10.1080/23273798.2019.1622750} {The use of context in resolving syntactic ambiguity: Structural and semantic influences}.
\newblock \emph{Language, cognition and neuroscience}, 35(1):43--57.

\bibitem[{Brown et~al.(2020)Brown, Mann, Ryder, Subbiah, Kaplan, Dhariwal, Neelakantan, Shyam, Sastry, Askell, Agarwal, Herbert-Voss, Krueger, Henighan, Child, Ramesh, Ziegler, Wu, Winter, Hesse, Chen, Sigler, Litwin, Gray, Chess, Clark, Berner, McCandlish, Radford, Sutskever, and Amodei}]{brown2020languagemodelsfewshotlearners}
Tom~B. Brown, Benjamin Mann, Nick Ryder, Melanie Subbiah, Jared Kaplan, Prafulla Dhariwal, Arvind Neelakantan, Pranav Shyam, Girish Sastry, Amanda Askell, Sandhini Agarwal, Ariel Herbert-Voss, Gretchen Krueger, Tom Henighan, Rewon Child, Aditya Ramesh, Daniel~M. Ziegler, Jeffrey Wu, Clemens Winter, and 12 others. 2020.
\newblock Language models are few-shot learners.
\newblock In \emph{Proceedings of the 34th International Conference on Neural Information Processing Systems}, NIPS '20, Red Hook, NY, USA. Curran Associates Inc.

\bibitem[{Cai et~al.(2024)Cai, Duan, Haslett, Wang, and Pickering}]{cai2024largelanguagemodelsresemble}
Zhenguang Cai, Xufeng Duan, David Haslett, Shuqi Wang, and Martin Pickering. 2024.
\newblock \href {https://doi.org/10.18653/v1/2024.cmcl-1.4} {Do large language models resemble humans in language use?}
\newblock In \emph{Proceedings of the Workshop on Cognitive Modeling and Computational Linguistics}, pages 37--56, Bangkok, Thailand. Association for Computational Linguistics.

\bibitem[{Chiang and Lee(2023)}]{chiang2023closerlookautomaticevaluation}
Cheng-Han Chiang and Hung-yi Lee. 2023.
\newblock \href {https://doi.org/10.18653/v1/2023.findings-emnlp.599} {A closer look into using large language models for automatic evaluation}.
\newblock In \emph{Findings of the Association for Computational Linguistics: EMNLP 2023}, pages 8928--8942, Singapore. Association for Computational Linguistics.

\bibitem[{Davis and van Schijndel(2020)}]{davis-van-schijndel-2020-discourse}
Forrest Davis and Marten van Schijndel. 2020.
\newblock \href {https://doi.org/10.18653/v1/2020.conll-1.32} {Discourse structure interacts with reference but not syntax in neural language models}.
\newblock In \emph{Proceedings of the 24th Conference on Computational Natural Language Learning}, pages 396--407, Online. Association for Computational Linguistics.

\bibitem[{Dubey et~al.(2024)Dubey, Jauhri, Pandey, Kadian, Al-Dahle, Letman, Mathur, Schelten, Yang, Fan, Goyal, Hartshorn, Yang, Mitra, Sravankumar, Korenev, Hinsvark, Rao, Zhang, Rodriguez, Gregerson, Spataru, Roziere, Biron, Tang, Chern, Caucheteux, Nayak, Bi, Marra, McConnell, Keller, Touret, Wu, Wong, Ferrer, Nikolaidis, Allonsius, Song, Pintz, Livshits, Esiobu, Choudhary, Mahajan, Garcia-Olano, Perino, Hupkes, Lakomkin, AlBadawy, Lobanova, Dinan, Smith, Radenovic, Zhang, Synnaeve, Lee, Anderson, Nail, Mialon, Pang, Cucurell, Nguyen, Korevaar, Xu, Touvron, Zarov, Ibarra, Kloumann, Misra, Evtimov, Copet, Lee, Geffert, Vranes, Park, Mahadeokar, Shah, van~der Linde, Billock, Hong, Lee, Fu, Chi, Huang, Liu, Wang, Yu, Bitton, Spisak, Park, Rocca, Johnstun, Saxe, Jia, Alwala, Upasani, Plawiak, Li, Heafield, Stone, El-Arini, Iyer, Malik, Chiu, Bhalla, Rantala-Yeary, van~der Maaten, Chen, Tan, Jenkins, Martin, Madaan, Malo, Blecher, Landzaat, de~Oliveira, Muzzi, Pasupuleti, Singh, Paluri, Kardas, Oldham, Rita,
  Pavlova, Kambadur, Lewis, Si, Singh, Hassan, Goyal, Torabi, Bashlykov, Bogoychev, Chatterji, Duchenne, Çelebi, Alrassy, Zhang, Li, Vasic, Weng, Bhargava, Dubal, Krishnan, Koura, Xu, He, Dong, Srinivasan, Ganapathy, Calderer, Cabral, Stojnic, Raileanu, Girdhar, Patel, Sauvestre, Polidoro, Sumbaly, Taylor, Silva, Hou, Wang, Hosseini, Chennabasappa, Singh, Bell, Kim, Edunov, Nie, Narang, Raparthy, Shen, Wan, Bhosale, Zhang, Vandenhende, Batra, Whitman, Sootla, Collot, Gururangan, Borodinsky, Herman, Fowler, Sheasha, Georgiou, Scialom, Speckbacher, Mihaylov, Xiao, Karn, Goswami, Gupta, Ramanathan, Kerkez, Gonguet, Do, Vogeti, Petrovic, Chu, Xiong, Fu, Meers, Martinet, Wang, Tan, Xie, Jia, Wang, Goldschlag, Gaur, Babaei, Wen, Song, Zhang, Li, Mao, Coudert, Yan, Chen, Papakipos, Singh, Grattafiori, Jain, Kelsey, Shajnfeld, Gangidi, Victoria, Goldstand, Menon, Sharma, Boesenberg, Vaughan, Baevski, Feinstein, Kallet, Sangani, Yunus, Lupu, Alvarado, Caples, Gu, Ho, Poulton, Ryan, Ramchandani, Franco, Saraf,
  Chowdhury, Gabriel, Bharambe, Eisenman, Yazdan, James, Maurer, Leonhardi, Huang, Loyd, Paola, Paranjape, Liu, Wu, Ni, Hancock, Wasti, Spence, Stojkovic, Gamido, Montalvo, Parker, Burton, Mejia, Wang, Kim, Zhou, Hu, Chu, Cai, Tindal, Feichtenhofer, Civin, Beaty, Kreymer, Li, Wyatt, Adkins, Xu, Testuggine, David, Parikh, Liskovich, Foss, Wang, Le, Holland, Dowling, Jamil, Montgomery, Presani, Hahn, Wood, Brinkman, Arcaute, Dunbar, Smothers, Sun, Kreuk, Tian, Ozgenel, Caggioni, Guzmán, Kanayet, Seide, Florez, Schwarz, Badeer, Swee, Halpern, Thattai, Herman, Sizov, Guangyi, Zhang, Lakshminarayanan, Shojanazeri, Zou, Wang, Zha, Habeeb, Rudolph, Suk, Aspegren, Goldman, Damlaj, Molybog, Tufanov, Veliche, Gat, Weissman, Geboski, Kohli, Asher, Gaya, Marcus, Tang, Chan, Zhen, Reizenstein, Teboul, Zhong, Jin, Yang, Cummings, Carvill, Shepard, McPhie, Torres, Ginsburg, Wang, Wu, U, Saxena, Prasad, Khandelwal, Zand, Matosich, Veeraraghavan, Michelena, Li, Huang, Chawla, Lakhotia, Huang, Chen, Garg, A, Silva, Bell,
  Zhang, Guo, Yu, Moshkovich, Wehrstedt, Khabsa, Avalani, Bhatt, Tsimpoukelli, Mankus, Hasson, Lennie, Reso, Groshev, Naumov, Lathi, Keneally, Seltzer, Valko, Restrepo, Patel, Vyatskov, Samvelyan, Clark, Macey, Wang, Hermoso, Metanat, Rastegari, Bansal, Santhanam, Parks, White, Bawa, Singhal, Egebo, Usunier, Laptev, Dong, Zhang, Cheng, Chernoguz, Hart, Salpekar, Kalinli, Kent, Parekh, Saab, Balaji, Rittner, Bontrager, Roux, Dollar, Zvyagina, Ratanchandani, Yuvraj, Liang, Alao, Rodriguez, Ayub, Murthy, Nayani, Mitra, Li, Hogan, Battey, Wang, Maheswari, Howes, Rinott, Bondu, Datta, Chugh, Hunt, Dhillon, Sidorov, Pan, Verma, Yamamoto, Ramaswamy, Lindsay, Lindsay, Feng, Lin, Zha, Shankar, Zhang, Zhang, Wang, Agarwal, Sajuyigbe, Chintala, Max, Chen, Kehoe, Satterfield, Govindaprasad, Gupta, Cho, Virk, Subramanian, Choudhury, Goldman, Remez, Glaser, Best, Kohler, Robinson, Li, Zhang, Matthews, Chou, Shaked, Vontimitta, Ajayi, Montanez, Mohan, Kumar, Mangla, Albiero, Ionescu, Poenaru, Mihailescu, Ivanov, Li, Wang,
  Jiang, Bouaziz, Constable, Tang, Wang, Wu, Wang, Xia, Wu, Gao, Chen, Hu, Jia, Qi, Li, Zhang, Zhang, Adi, Nam, Yu, Wang, Hao, Qian, He, Rait, DeVito, Rosnbrick, Wen, Yang, and Zhao}]{dubey2024llama3herdmodels}
Abhimanyu Dubey, Abhinav Jauhri, Abhinav Pandey, Abhishek Kadian, Ahmad Al-Dahle, Aiesha Letman, Akhil Mathur, Alan Schelten, Amy Yang, Angela Fan, Anirudh Goyal, Anthony Hartshorn, Aobo Yang, Archi Mitra, Archie Sravankumar, Artem Korenev, Arthur Hinsvark, Arun Rao, Aston Zhang, and 516 others. 2024.
\newblock \href {https://arxiv.org/abs/2407.21783} {The llama 3 herd of models}.
\newblock \emph{Preprint}, arXiv:2407.21783.

\bibitem[{Dutta et~al.(2020)Dutta, Wei, Yueksel, Chen, Liu, and Varshney}]{dutta-et-al-tradeoff}
Sanghamitra Dutta, Dennis Wei, Hazar Yueksel, Pin-Yu Chen, Sijia Liu, and Kush~R. Varshney. 2020.
\newblock Is there a trade-off between fairness and accuracy? a perspective using mismatched hypothesis testing.
\newblock In \emph{Proceedings of the 37th International Conference on Machine Learning}, ICML'20. JMLR.org.

\bibitem[{Ettinger(2020)}]{ettinger-2020-bert}
Allyson Ettinger. 2020.
\newblock \href {https://doi.org/10.1162/tacl_a_00298} {What {BERT} is not: Lessons from a new suite of psycholinguistic diagnostics for language models}.
\newblock \emph{Transactions of the Association for Computational Linguistics}, 8:34--48.

\bibitem[{Gan et~al.(2024)Gan, Poesio, and Yu}]{gan-etal-2024-assessing}
Yujian Gan, Massimo Poesio, and Juntao Yu. 2024.
\newblock \href {https://aclanthology.org/2024.lrec-main.145/} {Assessing the capabilities of large language models in coreference: An evaluation}.
\newblock In \emph{Proceedings of the 2024 Joint International Conference on Computational Linguistics, Language Resources and Evaluation (LREC-COLING 2024)}, pages 1645--1665, Torino, Italia. ELRA and ICCL.

\bibitem[{Ide et~al.(2025)Ide, Nishida, Vasselli, Oba, Sakai, Kamigaito, and Watanabe}]{ide2024makellmsgrammaticalknowledge}
Yusuke Ide, Yuto Nishida, Justin Vasselli, Miyu Oba, Yusuke Sakai, Hidetaka Kamigaito, and Taro Watanabe. 2025.
\newblock \href {https://aclanthology.org/2025.naacl-long.380/} {How to make the most of {LLM}s' grammatical knowledge for acceptability judgments}.
\newblock In \emph{Proceedings of the 2025 Conference of the Nations of the Americas Chapter of the Association for Computational Linguistics: Human Language Technologies (Volume 1: Long Papers)}, pages 7416--7432, Albuquerque, New Mexico. Association for Computational Linguistics.

\bibitem[{Kalai et~al.(2025)Kalai, Nachum, Vempala, and Zhang}]{kalai2025languagemodelshallucinate}
Adam~Tauman Kalai, Ofir Nachum, Santosh~S. Vempala, and Edwin Zhang. 2025.
\newblock \href {https://arxiv.org/abs/2509.04664} {Why language models hallucinate}.
\newblock \emph{Preprint}, arXiv:2509.04664.

\bibitem[{Kim et~al.(2023)Kim, Kim, Jeon, Park, and Kang}]{kim-etal-2023-tree}
Gangwoo Kim, Sungdong Kim, Byeongguk Jeon, Joonsuk Park, and Jaewoo Kang. 2023.
\newblock \href {https://doi.org/10.18653/v1/2023.emnlp-main.63} {Tree of clarifications: Answering ambiguous questions with retrieval-augmented large language models}.
\newblock In \emph{Proceedings of the 2023 Conference on Empirical Methods in Natural Language Processing}, pages 996--1009, Singapore. Association for Computational Linguistics.

\bibitem[{Kim et~al.(2024)Kim, Kim, Park, Kim, Park, Yoo, Lee, and Kim}]{kim-etal-2024-aligning}
Hyuhng~Joon Kim, Youna Kim, Cheonbok Park, Junyeob Kim, Choonghyun Park, Kang~Min Yoo, Sang-goo Lee, and Taeuk Kim. 2024.
\newblock \href {https://doi.org/10.18653/v1/2024.emnlp-main.119} {Aligning language models to explicitly handle ambiguity}.
\newblock In \emph{Proceedings of the 2024 Conference on Empirical Methods in Natural Language Processing}, pages 1989--2007, Miami, Florida, USA. Association for Computational Linguistics.

\bibitem[{Kishore and He(2024)}]{kishore-he-2024-unveiling}
Sindhu Kishore and Hangfeng He. 2024.
\newblock \href {https://doi.org/10.18653/v1/2024.naacl-short.20} {Unveiling divergent inductive biases of {LLM}s on temporal data}.
\newblock In \emph{Proceedings of the 2024 Conference of the North American Chapter of the Association for Computational Linguistics: Human Language Technologies (Volume 2: Short Papers)}, pages 220--228, Mexico City, Mexico. Association for Computational Linguistics.

\bibitem[{Lampinen et~al.(2024)Lampinen, Chan, Singh, and Shanahan}]{lampinen2024broaderspectrumincontextlearning}
Andrew~Kyle Lampinen, Stephanie C.~Y. Chan, Aaditya~K. Singh, and Murray Shanahan. 2024.
\newblock \href {https://arxiv.org/abs/2412.03782} {The broader spectrum of in-context learning}.
\newblock \emph{Preprint}, arXiv:2412.03782.

\bibitem[{Le and Ritter(2023)}]{le2023largelanguagemodelsrobust}
Nghia~T. Le and Alan Ritter. 2023.
\newblock \href {https://arxiv.org/abs/2305.14489} {Are large language models robust coreference resolvers?}
\newblock \emph{Preprint}, arXiv:2305.14489.

\bibitem[{Lee et~al.(2024)Lee, Scheinberg, Shore, and Agrawal}]{lee-etal-2024-multilingual}
So~Young Lee, Russell Scheinberg, Amber Shore, and Ameeta Agrawal. 2024.
\newblock \href {https://aclanthology.org/2024.paclic-1.42/} {Multilingual relative clause attachment ambiguity resolution in large language models}.
\newblock In \emph{Proceedings of the 38th Pacific Asia Conference on Language, Information and Computation}, pages 417--432, Tokyo, Japan. Tokyo University of Foreign Studies.

\bibitem[{Lee et~al.(2025)Lee, Scheinberg, Shore, and Agrawal}]{lee-etal-2025-relies}
So~Young Lee, Russell Scheinberg, Amber Shore, and Ameeta Agrawal. 2025.
\newblock \href {https://doi.org/10.18653/v1/2025.naacl-long.177} {Who relies more on world knowledge and bias for syntactic ambiguity resolution: Humans or {LLM}s?}
\newblock In \emph{Proceedings of the 2025 Conference of the Nations of the Americas Chapter of the Association for Computational Linguistics: Human Language Technologies (Volume 1: Long Papers)}, pages 3484--3498, Albuquerque, New Mexico. Association for Computational Linguistics.

\bibitem[{Levesque et~al.(2012)Levesque, Davis, and Morgenstern}]{levesque-wino-2012}
Hector~J. Levesque, Ernest Davis, and Leora Morgenstern. 2012.
\newblock The winograd schema challenge.
\newblock In \emph{Proceedings of the Thirteenth International Conference on Principles of Knowledge Representation and Reasoning}, KR'12, page 552–561. AAAI Press.

\bibitem[{Liu et~al.(2023{\natexlab{a}})Liu, Wu, Michael, Suhr, West, Koller, Swayamdipta, Smith, and Choi}]{liu-etal-2023-afraid}
Alisa Liu, Zhaofeng Wu, Julian Michael, Alane Suhr, Peter West, Alexander Koller, Swabha Swayamdipta, Noah Smith, and Yejin Choi. 2023{\natexlab{a}}.
\newblock \href {https://doi.org/10.18653/v1/2023.emnlp-main.51} {We{'}re afraid language models aren{'}t modeling ambiguity}.
\newblock In \emph{Proceedings of the 2023 Conference on Empirical Methods in Natural Language Processing}, pages 790--807, Singapore. Association for Computational Linguistics.

\bibitem[{Liu et~al.(2023{\natexlab{b}})Liu, Mao, Luu, and Cambria}]{liu2023surveycorefresolution}
Ruicheng Liu, Rui Mao, Anh~Tuan Luu, and Erik Cambria. 2023{\natexlab{b}}.
\newblock \href {https://doi.org/10.1007/s10462-023-10506-3} {A brief survey on recent advances in coreference resolution}.
\newblock \emph{Artif. Intell. Rev.}, 56(12):14439–14481.

\bibitem[{Liu et~al.(2025)Liu, Peng, Cao, Bo, Shen, Du, Cheng, Wang, Yin, and Zhang}]{liu2025bridgingcontextgapsleveraging}
Yanming Liu, Xinyue Peng, Jiannan Cao, Shi Bo, Yanxin Shen, Tianyu Du, Sheng Cheng, Xun Wang, Jianwei Yin, and Xuhong Zhang. 2025.
\newblock \href {https://arxiv.org/abs/2410.01671} {Bridging context gaps: Leveraging coreference resolution for long contextual understanding}.
\newblock \emph{Preprint}, arXiv:2410.01671.

\bibitem[{Lucy et~al.(2024)Lucy, Blodgett, Shokouhi, Wallach, and Olteanu}]{lucy-etal-2024-system-behaviors}
Li~Lucy, Su~Lin Blodgett, Milad Shokouhi, Hanna Wallach, and Alexandra Olteanu. 2024.
\newblock \href {https://doi.org/10.18653/v1/2024.naacl-long.61} {{\textquotedblleft}one-size-fits-all{\textquotedblright}? examining expectations around what constitute {\textquotedblleft}fair{\textquotedblright} or {\textquotedblleft}good{\textquotedblright} {NLG} system behaviors}.
\newblock In \emph{Proceedings of the 2024 Conference of the North American Chapter of the Association for Computational Linguistics: Human Language Technologies (Volume 1: Long Papers)}, pages 1054--1089, Mexico City, Mexico. Association for Computational Linguistics.

\bibitem[{Niwa and Iso(2024)}]{niwa2024ambignlgaddressingtaskambiguity}
Ayana Niwa and Hayate Iso. 2024.
\newblock \href {https://doi.org/10.18653/v1/2024.emnlp-main.599} {{A}mbig{NLG}: Addressing task ambiguity in instruction for {NLG}}.
\newblock In \emph{Proceedings of the 2024 Conference on Empirical Methods in Natural Language Processing}, pages 10733--10752, Miami, Florida, USA. Association for Computational Linguistics.

\bibitem[{OpenAI(2025)}]{openai-o3-2025}
OpenAI. 2025.
\newblock \href {https://openai.com/index/o3-o4-mini-system-card/} {Openai o3 and o4-mini system card}.

\bibitem[{OpenAI et~al.(2024)OpenAI, Achiam, Adler, Agarwal, Ahmad, Akkaya, Aleman, Almeida, Altenschmidt, Altman, Anadkat, Avila, Babuschkin, Balaji, Balcom, Baltescu, Bao, Bavarian, Belgum, Bello, Berdine, Bernadett-Shapiro, Berner, Bogdonoff, Boiko, Boyd, Brakman, Brockman, Brooks, Brundage, Button, Cai, Campbell, Cann, Carey, Carlson, Carmichael, Chan, Chang, Chantzis, Chen, Chen, Chen, Chen, Chen, Chess, Cho, Chu, Chung, Cummings, Currier, Dai, Decareaux, Degry, Deutsch, Deville, Dhar, Dohan, Dowling, Dunning, Ecoffet, Eleti, Eloundou, Farhi, Fedus, Felix, Fishman, Forte, Fulford, Gao, Georges, Gibson, Goel, Gogineni, Goh, Gontijo-Lopes, Gordon, Grafstein, Gray, Greene, Gross, Gu, Guo, Hallacy, Han, Harris, He, Heaton, Heidecke, Hesse, Hickey, Hickey, Hoeschele, Houghton, Hsu, Hu, Hu, Huizinga, Jain, Jain, Jang, Jiang, Jiang, Jin, Jin, Jomoto, Jonn, Jun, Kaftan, Łukasz Kaiser, Kamali, Kanitscheider, Keskar, Khan, Kilpatrick, Kim, Kim, Kim, Kirchner, Kiros, Knight, Kokotajlo, Łukasz Kondraciuk,
  Kondrich, Konstantinidis, Kosic, Krueger, Kuo, Lampe, Lan, Lee, Leike, Leung, Levy, Li, Lim, Lin, Lin, Litwin, Lopez, Lowe, Lue, Makanju, Malfacini, Manning, Markov, Markovski, Martin, Mayer, Mayne, McGrew, McKinney, McLeavey, McMillan, McNeil, Medina, Mehta, Menick, Metz, Mishchenko, Mishkin, Monaco, Morikawa, Mossing, Mu, Murati, Murk, Mély, Nair, Nakano, Nayak, Neelakantan, Ngo, Noh, Ouyang, O'Keefe, Pachocki, Paino, Palermo, Pantuliano, Parascandolo, Parish, Parparita, Passos, Pavlov, Peng, Perelman, de~Avila Belbute~Peres, Petrov, de~Oliveira~Pinto, Michael, Pokorny, Pokrass, Pong, Powell, Power, Power, Proehl, Puri, Radford, Rae, Ramesh, Raymond, Real, Rimbach, Ross, Rotsted, Roussez, Ryder, Saltarelli, Sanders, Santurkar, Sastry, Schmidt, Schnurr, Schulman, Selsam, Sheppard, Sherbakov, Shieh, Shoker, Shyam, Sidor, Sigler, Simens, Sitkin, Slama, Sohl, Sokolowsky, Song, Staudacher, Such, Summers, Sutskever, Tang, Tezak, Thompson, Tillet, Tootoonchian, Tseng, Tuggle, Turley, Tworek, Uribe, Vallone,
  Vijayvergiya, Voss, Wainwright, Wang, Wang, Wang, Ward, Wei, Weinmann, Welihinda, Welinder, Weng, Weng, Wiethoff, Willner, Winter, Wolrich, Wong, Workman, Wu, Wu, Wu, Xiao, Xu, Yoo, Yu, Yuan, Zaremba, Zellers, Zhang, Zhang, Zhao, Zheng, Zhuang, Zhuk, and Zoph}]{openai2024gpt4technicalreport}
OpenAI, Josh Achiam, Steven Adler, Sandhini Agarwal, Lama Ahmad, Ilge Akkaya, Florencia~Leoni Aleman, Diogo Almeida, Janko Altenschmidt, Sam Altman, Shyamal Anadkat, Red Avila, Igor Babuschkin, Suchir Balaji, Valerie Balcom, Paul Baltescu, Haiming Bao, Mohammad Bavarian, Jeff Belgum, and 262 others. 2024.
\newblock \href {https://arxiv.org/abs/2303.08774} {Gpt-4 technical report}.
\newblock \emph{Preprint}, arXiv:2303.08774.

\bibitem[{Ortega-Martín et~al.(2023)Ortega-Martín, Óscar García-Sierra, Ardoiz, Álvarez, Armenteros, and Alonso}]{ortegamartín2023linguisticambiguityanalysischatgpt}
Miguel Ortega-Martín, Óscar García-Sierra, Alfonso Ardoiz, Jorge Álvarez, Juan~Carlos Armenteros, and Adrián Alonso. 2023.
\newblock \href {https://arxiv.org/abs/2302.06426} {Linguistic ambiguity analysis in chatgpt}.
\newblock \emph{Preprint}, arXiv:2302.06426.

\bibitem[{Pezeshkpour and Hruschka(2024)}]{pezeshkpour2023largelanguagemodelssensitivity}
Pouya Pezeshkpour and Estevam Hruschka. 2024.
\newblock \href {https://doi.org/10.18653/v1/2024.findings-naacl.130} {Large language models sensitivity to the order of options in multiple-choice questions}.
\newblock In \emph{Findings of the Association for Computational Linguistics: NAACL 2024}, pages 2006--2017, Mexico City, Mexico. Association for Computational Linguistics.

\bibitem[{Pradhan et~al.(2012)Pradhan, Moschitti, Xue, Uryupina, and Zhang}]{pradhan-etal-2012-conll}
Sameer Pradhan, Alessandro Moschitti, Nianwen Xue, Olga Uryupina, and Yuchen Zhang. 2012.
\newblock \href {https://aclanthology.org/W12-4501/} {{C}o{NLL}-2012 shared task: Modeling multilingual unrestricted coreference in {O}nto{N}otes}.
\newblock In \emph{Joint Conference on {EMNLP} and {C}o{NLL} - Shared Task}, pages 1--40, Jeju Island, Korea. Association for Computational Linguistics.

\bibitem[{Qamar et~al.(2024)Qamar, Yasmeen, Pathak, Sohail, Øivind Madsen, and Rangarajan}]{qamar2024bigclaimslowoutcomes}
Md.~Tauseef Qamar, Juhi Yasmeen, Sanket~Kumar Pathak, Shahab~Saquib Sohail, Dag Øivind Madsen, and Mithila Rangarajan. 2024.
\newblock \href {https://doi.org/10.1080/23311983.2024.2353984} {Big claims, low outcomes: fact checking chatgpt’s efficacy in handling linguistic creativity and ambiguity}.
\newblock \emph{Cogent Arts \& Humanities}, 11(1):2353984.

\bibitem[{Qiu et~al.(2024)Qiu, Zhao, Ziser, Korhonen, Ponti, and Cohen}]{qiu-etal-2024-large}
Yifu Qiu, Zheng Zhao, Yftah Ziser, Anna Korhonen, Edoardo Ponti, and Shay Cohen. 2024.
\newblock \href {https://doi.org/10.18653/v1/2024.naacl-long.391} {Are large language model temporally grounded?}
\newblock In \emph{Proceedings of the 2024 Conference of the North American Chapter of the Association for Computational Linguistics: Human Language Technologies (Volume 1: Long Papers)}, pages 7064--7083, Mexico City, Mexico. Association for Computational Linguistics.

\bibitem[{Rahmani et~al.(2023)Rahmani, Wang, Feng, Zhang, Yilmaz, and Lipani}]{rahmani2023surveyaskingclarificationquestions}
Hossein~A. Rahmani, Xi~Wang, Yue Feng, Qiang Zhang, Emine Yilmaz, and Aldo Lipani. 2023.
\newblock \href {https://doi.org/10.18653/v1/2023.acl-long.152} {A survey on asking clarification questions datasets in conversational systems}.
\newblock In \emph{Proceedings of the 61st Annual Meeting of the Association for Computational Linguistics (Volume 1: Long Papers)}, pages 2698--2716, Toronto, Canada. Association for Computational Linguistics.

\bibitem[{Scheinberg et~al.(2025)Scheinberg, Lee, and Agrawal}]{Scheinberg2025MissingTC}
Russell Scheinberg, So~Young Lee, and Ameeta Agrawal. 2025.
\newblock \href {https://api.semanticscholar.org/CorpusID:278079654} {Missing the cues: Llms’ insensitivity to semantic biases in relative clause attachment}.
\newblock \emph{Proceedings of the Linguistic Society of America}.

\bibitem[{Seminck and Amsili(2017)}]{seminck-amsili-2017-computational}
Olga Seminck and Pascal Amsili. 2017.
\newblock \href {https://aclanthology.org/E17-4006} {A computational model of human preferences for pronoun resolution}.
\newblock In \emph{Proceedings of the Student Research Workshop at the 15th Conference of the {E}uropean Chapter of the Association for Computational Linguistics}, pages 53--63, Valencia, Spain. Association for Computational Linguistics.

\bibitem[{Skantze and Do{\u{g}}ru{\"o}z(2023)}]{skantze-dogruoz-2023-open}
Gabriel Skantze and A.~Seza Do{\u{g}}ru{\"o}z. 2023.
\newblock \href {https://doi.org/10.18653/v1/2023.sigdial-1.57} {The open-domain paradox for chatbots: Common ground as the basis for human-like dialogue}.
\newblock In \emph{Proceedings of the 24th Annual Meeting of the Special Interest Group on Discourse and Dialogue}, pages 605--614, Prague, Czechia. Association for Computational Linguistics.

\bibitem[{Winograd(1972)}]{WINOGRAD19721}
Terry Winograd. 1972.
\newblock \href {https://doi.org/10.1016/0010-0285(72)90002-3} {Understanding natural language}.
\newblock \emph{Cognitive Psychology}, 3(1):1--191.

\bibitem[{Yuan et~al.(2023)Yuan, Malaviya, and Yatskar}]{yuan2023ambicorefevaluatinghumanmodel}
Yuewei Yuan, Chaitanya Malaviya, and Mark Yatskar. 2023.
\newblock \href {https://doi.org/10.18653/v1/2023.findings-eacl.75} {{A}mbi{C}oref: Evaluating human and model sensitivity to ambiguous coreference}.
\newblock In \emph{Findings of the Association for Computational Linguistics: EACL 2023}, pages 1023--1030, Dubrovnik, Croatia. Association for Computational Linguistics.

\bibitem[{Zhang and Choi(2025)}]{zhang2023clarifynecessaryresolvingambiguity}
Michael~JQ Zhang and Eunsol Choi. 2025.
\newblock \href {https://aclanthology.org/2025.findings-naacl.306/} {Clarify when necessary: Resolving ambiguity through interaction with {LM}s}.
\newblock In \emph{Findings of the Association for Computational Linguistics: NAACL 2025}, pages 5526--5543, Albuquerque, New Mexico. Association for Computational Linguistics.

\bibitem[{Zhang et~al.(2024)Zhang, Qin, Deng, Huang, Lei, Liu, Jin, Liang, and Chua}]{zhang-etal-2024-clamber}
Tong Zhang, Peixin Qin, Yang Deng, Chen Huang, Wenqiang Lei, Junhong Liu, Dingnan Jin, Hongru Liang, and Tat-Seng Chua. 2024.
\newblock \href {https://doi.org/10.18653/v1/2024.acl-long.578} {{CLAMBER}: A benchmark of identifying and clarifying ambiguous information needs in large language models}.
\newblock In \emph{Proceedings of the 62nd Annual Meeting of the Association for Computational Linguistics (Volume 1: Long Papers)}, pages 10746--10766, Bangkok, Thailand. Association for Computational Linguistics.

\end{thebibliography}

\appendix

\section{In-depth Model Patterns}
\label{app:model_patterns}

\subsection{GPT-4o in-depth}
What explains GPT-4o's preference for A in the TOP category in both unambiguous and ambiguous contexts? The TOP sentences all use a phrase that begins with either "before" or "after" as their main structuring feature that contains the constraint in the unambiguous case and that lacks one in the ambiguous case. 
Sentences involving this before/after element require temporal processing, in this case combing with the appropriate reasoning about who could have participated in one event before or after another happened. GPT-4 has been shown to have a preference for "before" and "false" in textual entailment contexts \cite{kishore-he-2024-unveiling, qiu-etal-2024-large}. 

GPT-4o's ECO responses keep a similar pattern in both the unambiguous and ambiguous cases. In these sentences, the pronoun to be resolved is the subject position of the subordinate clause ("Matthew told Joshua that \textbf{he} offended the client." \textit{ambiguous}).
Compare "William told Joshua that he explained to the saleswoman." \textit{unambiguous}. Since William is the subject of the sentence, and the pronoun is the subject of the subordinate clause, this may be a strong enough link that the model doesn't have a reason to change its preferences in the ambiguous case. It is enough to assume that the subject, in English the more active entity in a sentence, goes on being the more active entity. 

While the ECO sentences place the pronoun in question at the subject position of the subordinate clause, The ECS sentences place it in the object position: "Melissa told Jennifer that the father-in-law terrified her." \textit{unambiguous} and "The sister-in-law told Amanda that the client envied her." \textit{ambiguous}. If the model is relying on a constant subject to disambiguate the ECO case, it cannot do so here. We see a pattern that agrees with this: in the ambiguous ECS sentences, the model shifts its preference to a more flat distribution of possible answers. See \ref{sec:ECO_ECS} for a discussion on these two categories.

We see similar behavior in the IC category, with the model retaining a preference for the \textbf{A/A?} answers in the ambiguous case. It does show a slight shift toward more uncertainty. The use of implied causality in these sentences means the model must draw on reasoning abilities in order to resolve the coreferent, and this may explain the significant difference we see between models, where in the unambiguous case GPT-4o is best at IC.

\subsection{Llama in-depth}
Llama 3.1 performs abysmally in the main task of resolving the coreferent in the unambiguous case. How it fails, however, is interesting. It vastly prefers the \textbf{?} (`mostly ambiguous') response in every case except for the IC category. While over the three runs it applies the ambiguous response to the ECO category the most frequently (at 445 total), it most consistently applies the ambiguous response to questions from sentences in the IC category (164 total). Each question occurs three times to match the three times its sentence occurs, but some questions are the same for different sentences due to the templates used to create them. For example, the "Who was looking for suggestions?" question pairs with 18 different sentences, and thus the model responds to it 54 times. It uses the ambiguous response to this question 24 times, almost half the times it is presented. This is a sentence structure that the model deals particularly poorly with.

\subsection{ECO vs. ECS}
\label{sec:ECO_ECS}

The two Experiencer-Constrained categories, ECO and ECS, should in theory have very similar results for all these measures. They are both constrained in the unambiguous case by the choice of verb in the subordinate clause, and differ from the ambiguous case only by verb choice. However, in all measures we see surprising differences in model behavior between these two categories.

\begin{enumerate}
    \item Accuracy in \textsc{Reflect} by model: Llama 3.1 does worst in ECS, while ECS is GPT-4o's second-best category (Table \ref{tab:baselines}).
    \item Scores under other paradigms can vary widely between the two -- see for example the abnormally high ECS score for WSC prompt in Llama 3.1 (Table \ref{tab:baselines}).
    \item Answer consistency: Humans have similar consistency on ECO and ECS, but models have about a 20\% difference in consistency rates for the two categories (Table \ref{tab:unanimous_answer}).
\end{enumerate}

In ECO, the pronoun to be resolved is located in the subject position of the subordinate clause, whereas in ECS it is in the object position. This could show a bias toward syntactic position, or subject preference, that benefits the ECO resolution and harms the ECS resolution tasks.

{We report the by-category chi-square test for statistical significance for these results in Table \ref{tab:stat_sig}. The $p$-values are all very small, showing that the answer patterns exhibited by the models for both unambiguous and ambiguous subsets are significantly different from the human answer patterns.}

\begin{table*}[!t]
\small
\setlength{\tabcolsep}{10pt}  
\begin{tabular}{lcccc|cccc}
\toprule
\multirow{3}{*}{} & \multicolumn{4}{c|}{\bf Unambiguous} & \multicolumn{4}{c}{\bf Ambiguous} \\
\cmidrule{2-9}
& \multicolumn{2}{c}{\bf Llama 3.1} & \multicolumn{2}{c}{\bf GPT-4o} & \multicolumn{2}{c}{\bf Llama 3.1} & \multicolumn{2}{c}{\bf GPT-4o}\\
& $\chi^2$  & $p$  & $\chi^2$ & $p$  & 
$\chi^2$  & $p$  & $\chi^2$  & $p$  \\
\midrule
ECO & 489.4 & 1.32e-104 & 44.5 & 5.10e-09 & 528.7 & 4.07e-113 & 492.8 & 2.37e-105\\
ECS  & 265.3 & 3.27e-56 & 14.0 & 7.18e-03 & 189.6 & 6.38e-40 & 80.7 & 1.26e-16 \\
IC & 218.7 & 3.59e-46 & 154.7 & 2.01e-32 & 536.3 & 9.32e-115 & 56.3 & 1.7e-11 \\
TOP & 356.8 & 5.99e-76 & 21.5 & 2.52e-04 & 475.8 & 1.13e-101 & 40.7 & 3.03e-08 \\
\bottomrule
\end{tabular}
\caption{Statistical significance test results for REFLECT experiments.}
\label{tab:stat_sig}
\end{table*}

\section{Results With Expanded Ambiguity Labels}
\label{app:fullanswers}
\begin{figure*}[h]
    \centering
    \begin{minipage}[b]{0.24\linewidth}
        \centering
        \includegraphics[width=\linewidth]{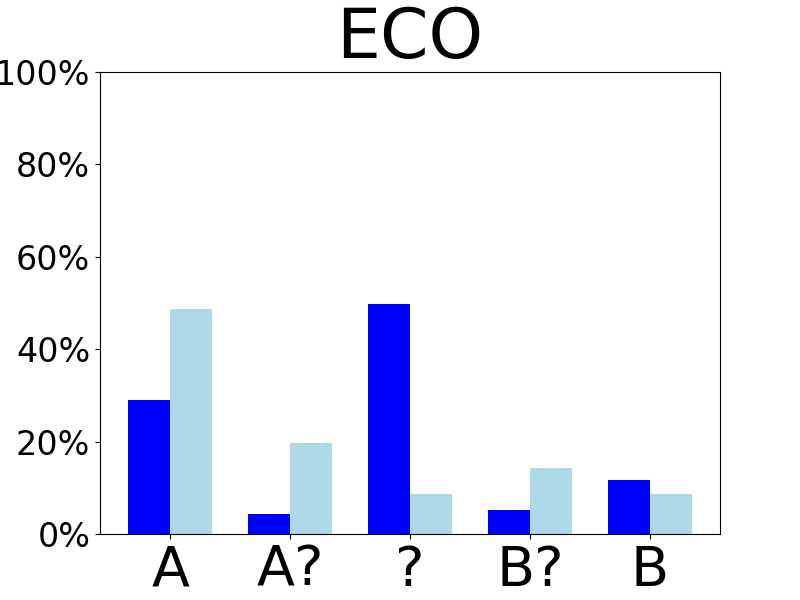}
    \end{minipage}
    \hfill
    \begin{minipage}[b]{0.24\linewidth}
        \centering
        \includegraphics[width=\linewidth]{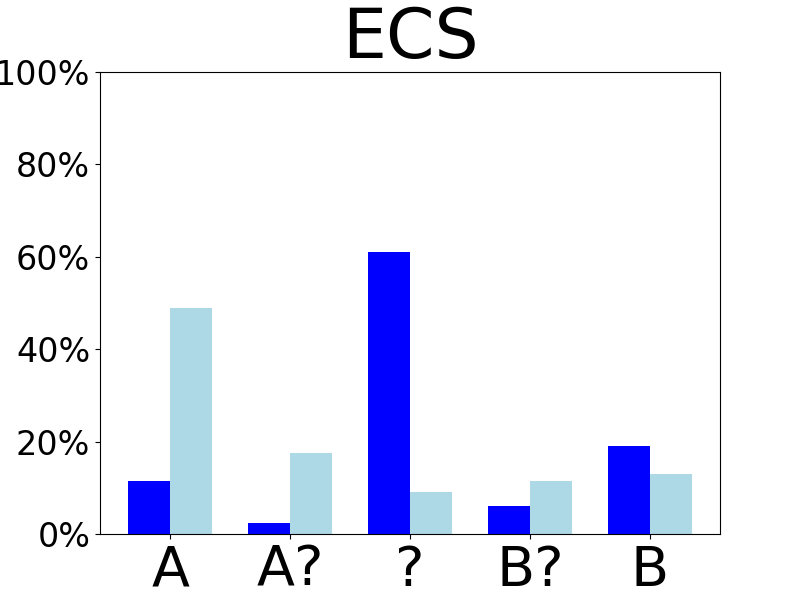}
    \end{minipage}
    \hfill
    \begin{minipage}[b]{0.24\linewidth}
        \centering
        \includegraphics[width=\linewidth]{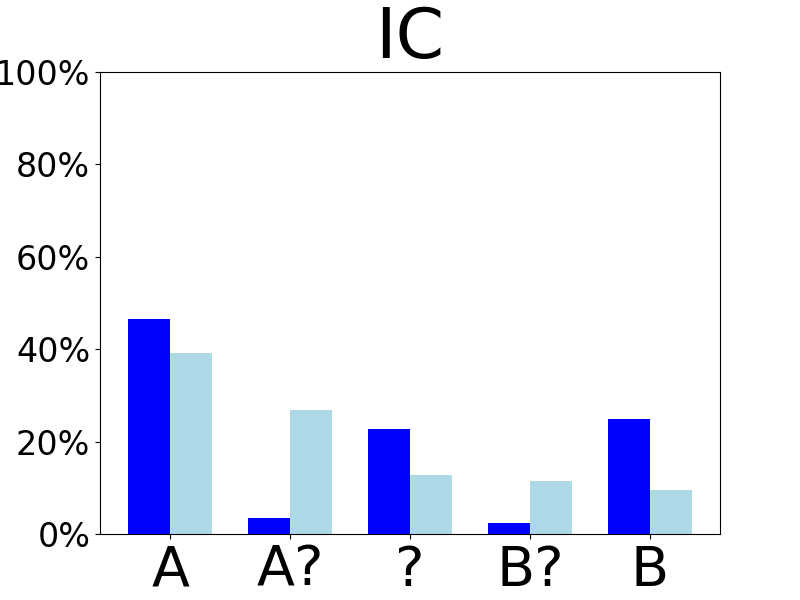}
    \end{minipage}
    \hfill
    \begin{minipage}[b]{0.24\linewidth}
        \centering
        \includegraphics[width=\linewidth]{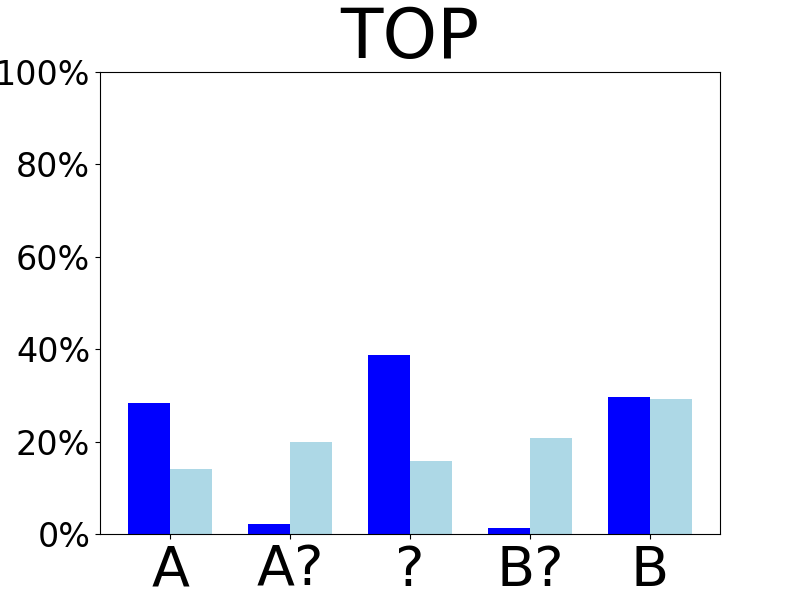}
    \end{minipage}
    \caption{\textbf{\textsc{Reflect}: Unambiguous Llama 3.1 70b results.} Dark blue are model results, light blue are averaged human judgments.}
    \label{fig:full_unamb_llama1}
\end{figure*}

\begin{figure*}[t!]
    \centering
    \begin{minipage}[b]{0.24\linewidth}
        \centering
        \includegraphics[width=\linewidth]{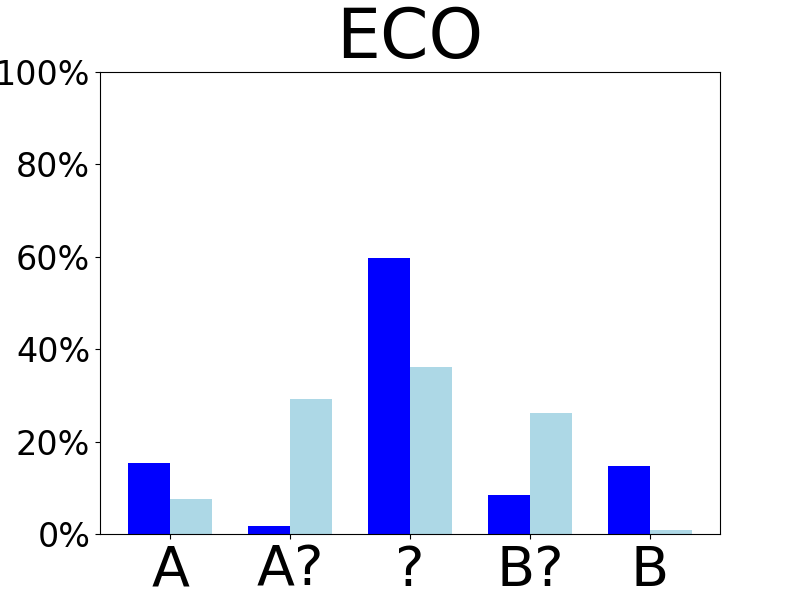}
    \end{minipage}
    \hfill
    \begin{minipage}[b]{0.24\linewidth}
        \centering
        \includegraphics[width=\linewidth]{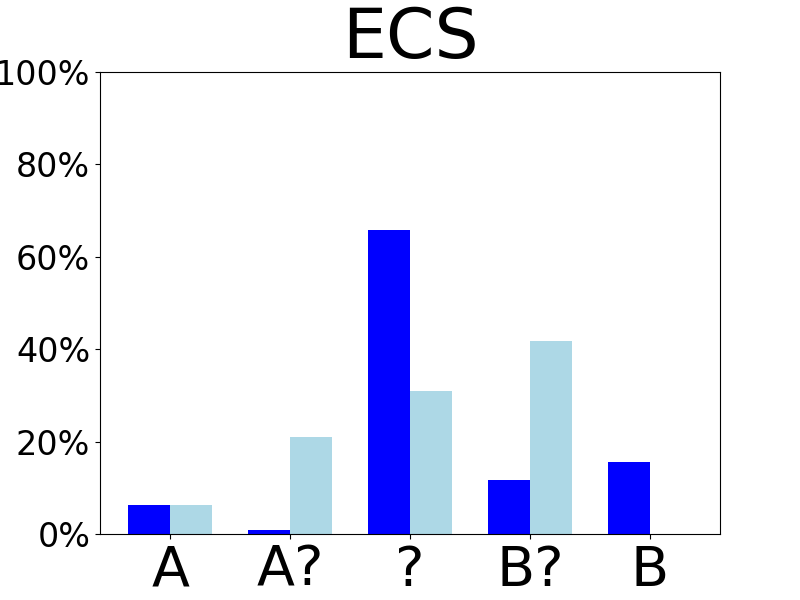}
    \end{minipage}
    \hfill
    \begin{minipage}[b]{0.24\linewidth}
        \centering
        \includegraphics[width=\linewidth]{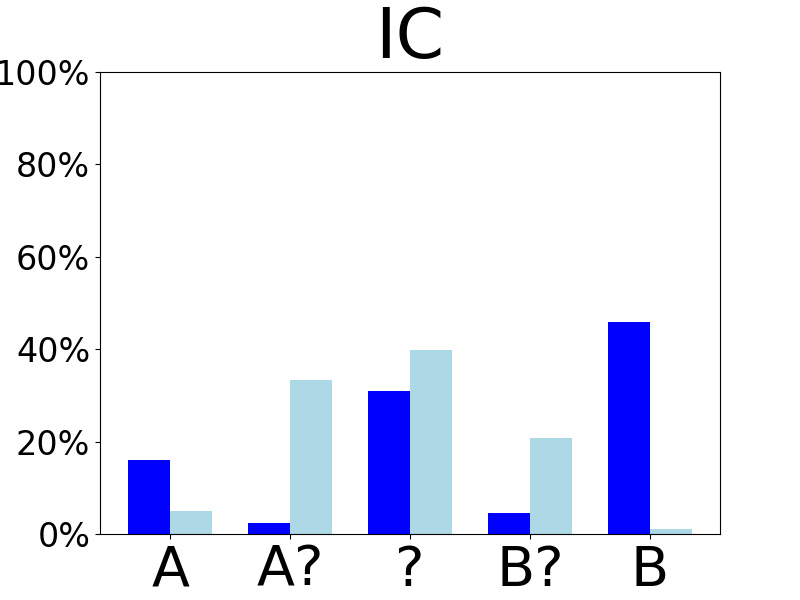}
    \end{minipage}
    \hfill
    \begin{minipage}[b]{0.24\linewidth}
        \centering
        \includegraphics[width=\linewidth]{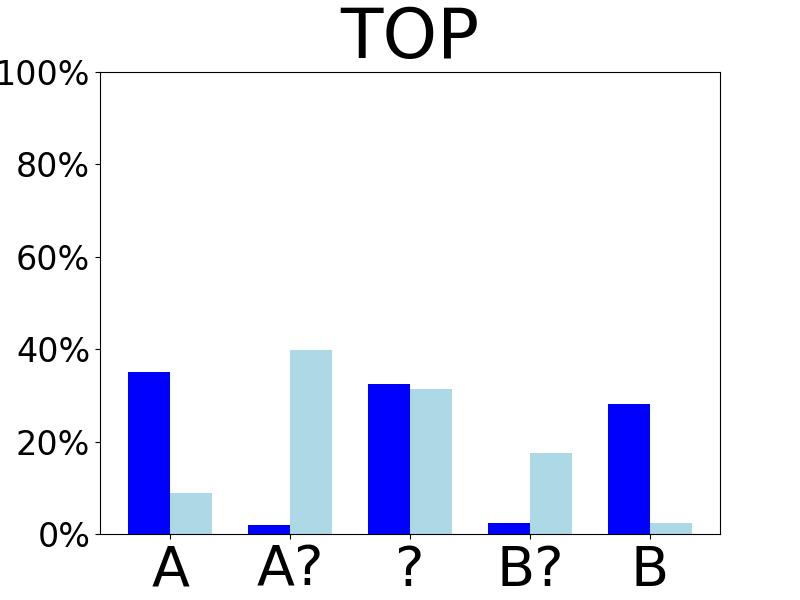}
    \end{minipage}
    \caption{\textbf{\textsc{Reflect}: Ambiguous Llama 3.1 70b results.} Dark blue are model results, light blue are averaged human judgments.}
    \label{fig:full_amb_llama1}
\end{figure*}

\begin{figure*}[!t]
    \centering
    \begin{minipage}[b]{0.24\linewidth}
        \centering
        \includegraphics[width=\linewidth]{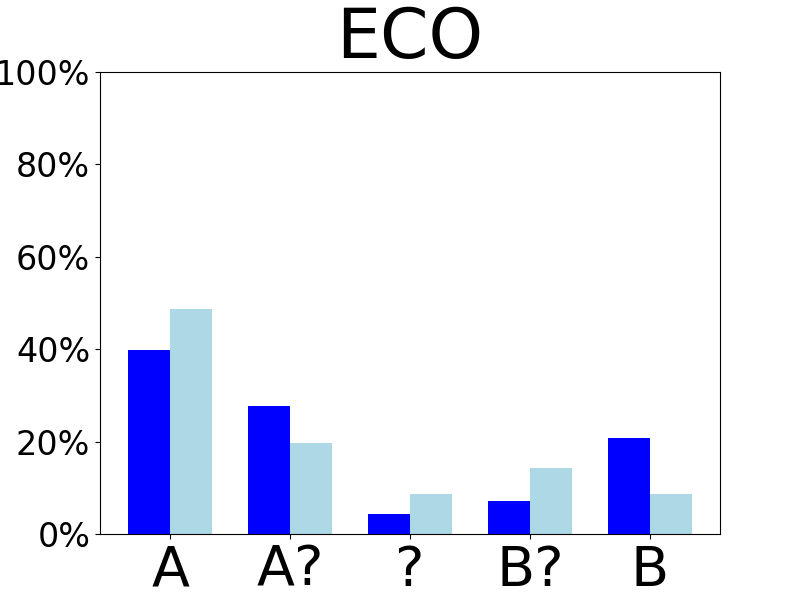}
    \end{minipage}
    \hfill
    \begin{minipage}[b]{0.24\linewidth}
        \centering
        \includegraphics[width=\linewidth]{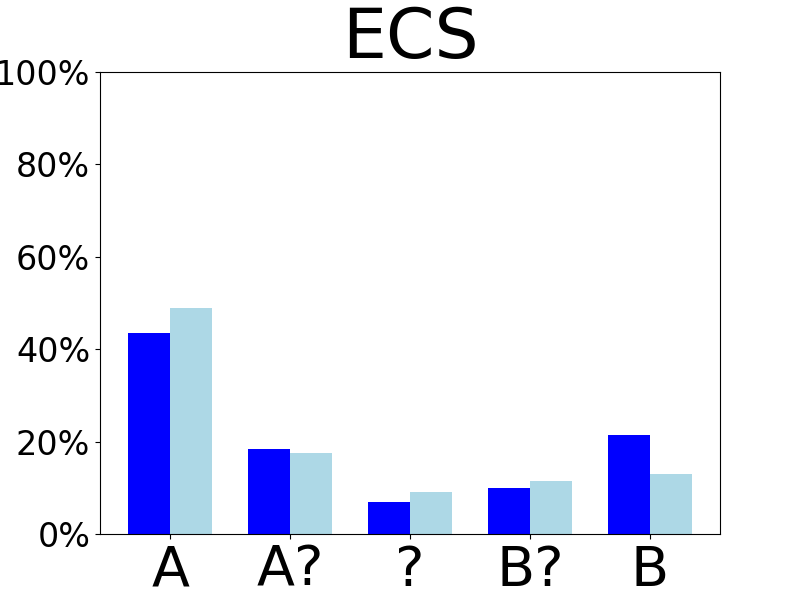}
    \end{minipage}
    \hfill
    \begin{minipage}[b]{0.24\linewidth}
        \centering
        \includegraphics[width=\linewidth]{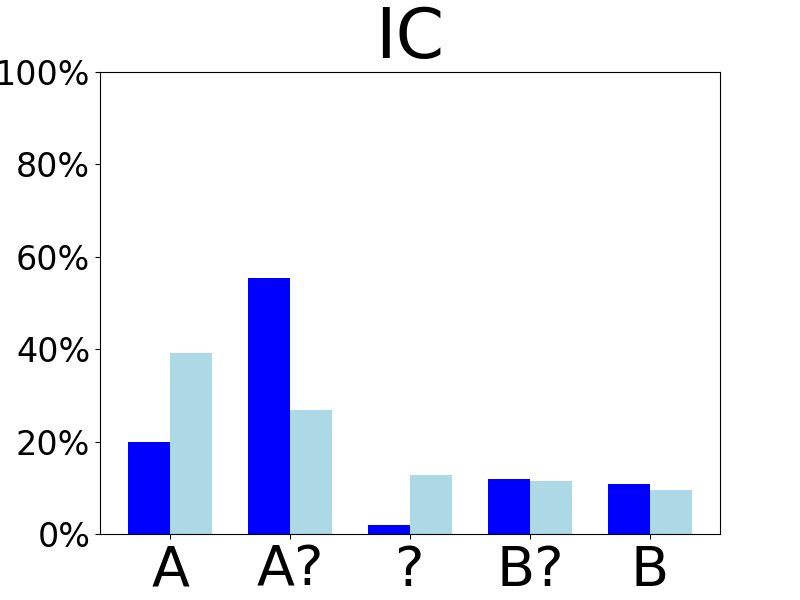}
    \end{minipage}
    \hfill
    \begin{minipage}[b]{0.24\linewidth}
        \centering
        \includegraphics[width=\linewidth]{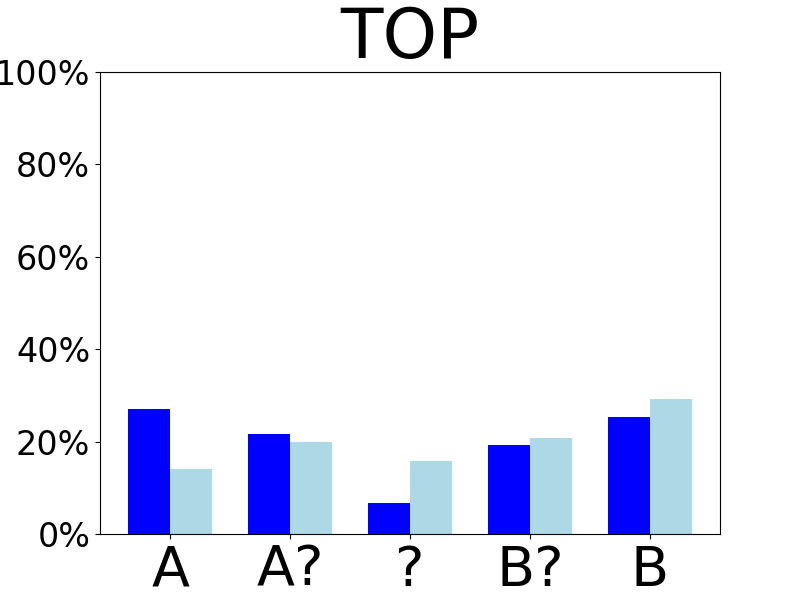}
    \end{minipage}
    \caption{\textbf{\textsc{Reflect}: Unambiguous GPT-4o results.} Dark blue are model results, light blue are averaged human judgments.}
    \label{fig:full_unamb_gpt1}
\end{figure*}

\begin{figure*}[!t]
    \centering
    \begin{minipage}[b]{0.24\linewidth}
        \centering
        \includegraphics[width=\linewidth]{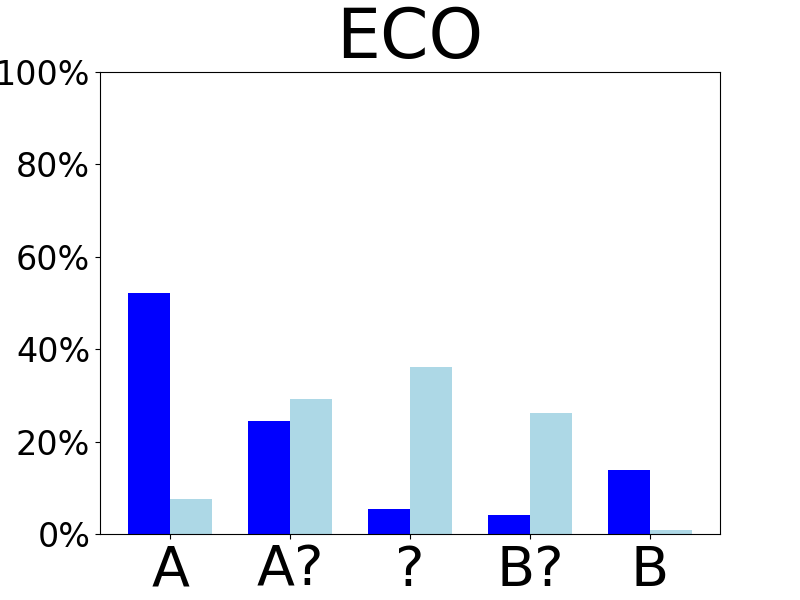}
    \end{minipage}
    \hfill
    \begin{minipage}[b]{0.24\linewidth}
        \centering
        \includegraphics[width=\linewidth]{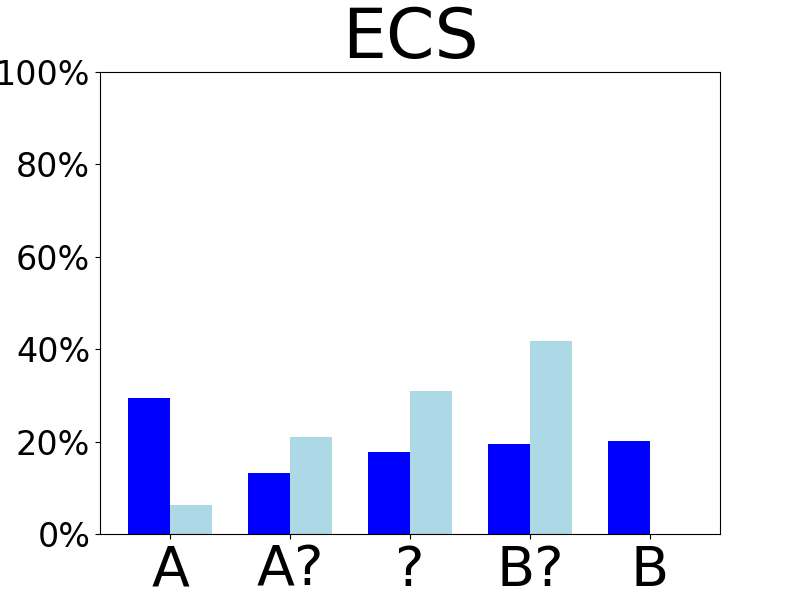}
    \end{minipage}
    \hfill
    \begin{minipage}[b]{0.24\linewidth}
        \centering
        \includegraphics[width=\linewidth]{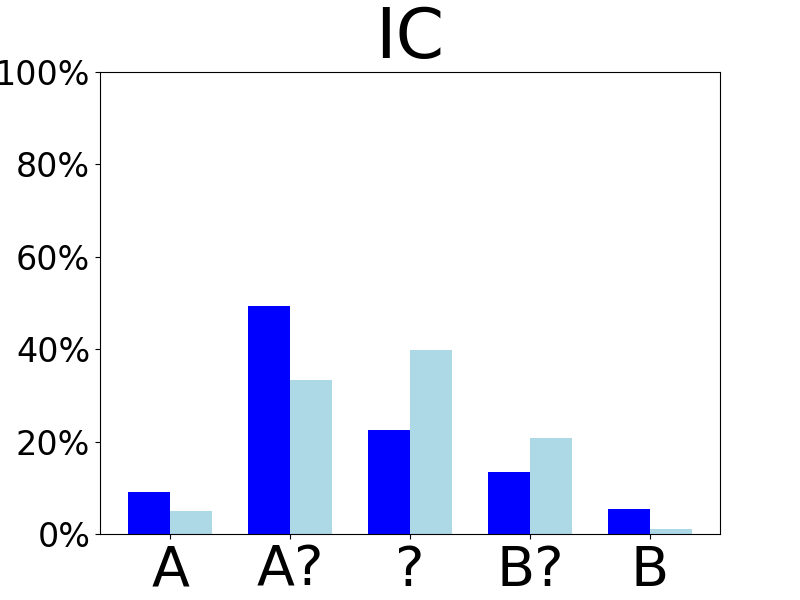}
    \end{minipage}
    \hfill
    \begin{minipage}[b]{0.24\linewidth}
        \centering
        \includegraphics[width=\linewidth]{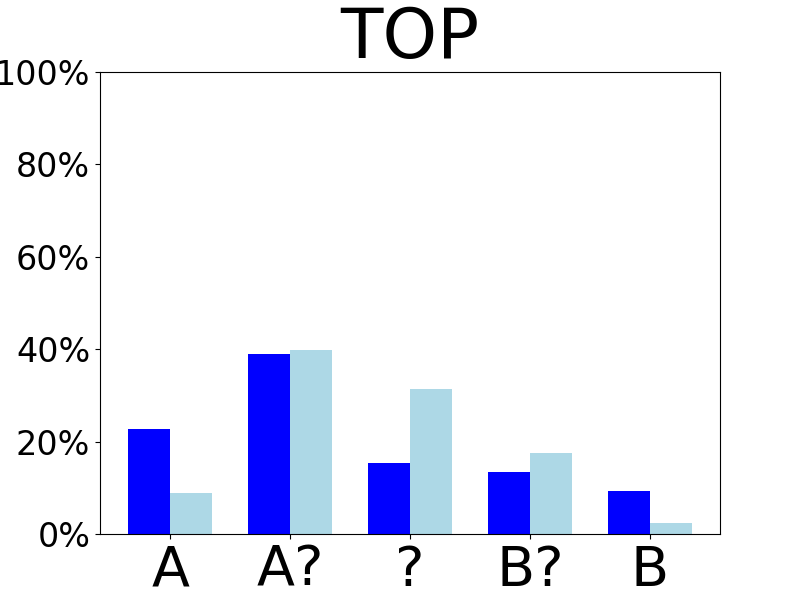}
    \end{minipage}
    \caption{\textbf{\textsc{Reflect}: Ambiguous GPT-4o results.} Dark blue are model results, light blue are averaged human judgments.}
    \label{fig:full_amb_gpt1}
\end{figure*}

Figures~\ref{fig:full_unamb_llama1}, \ref{fig:full_amb_llama1}, \ref{fig:full_unamb_gpt1}, and \ref{fig:full_amb_gpt1} present the full set of results using the \textsc{Reflect} prompt across all four categories.
{In Figure~\ref{fig:full_amb_gpt1}, } 
GPT-4o shows a pattern shift between unambiguous and ambiguous in the ECS, IC, and TOP categories. 
The shift in ECS does not match the shift in human preferences exactly, 
but it levels out the answers to show a variety of responses, 
instead of retaining the \textbf{A}-dominant focus from the unambiguous case. 
In the IC category, there is a notable shift toward the more uncertain responses, 
aligning more closely with the human responses while retaining some of the 
\textbf{A}-directed focus. 
For the TOP category uncertain answers increase, though the model shows a marked increase in preference for \textbf{A?}. In {Figure \ref{fig:full_amb_llama1}} for Llama 3.1 however, the model shifts clearly only in IC, with minor adjustments in other categories.


\section{Category Differences in \textit{Explain} Responses}
\label{app:explain}
In Table~\ref{tab:explain_correct_gpt}, we observe that for GPT-4o, the proportion of \textit{explain} responses  are higher in the categories where the model has low overall accuracy: 22.82\% \textit{explain} vs 30.26\% correct in {IC}, and 18.06\% \textit{explain} vs 35\% correct in {TOP}, compared to the much lower rates of around 7.5\% \textit{explain} and around 50\% correct in the other two categories.

\begin{table}[!ht]
  \centering
  \setlength{\tabcolsep}{6pt}
  \small
  \begin{tabular}{lccc}
    \toprule
    \textbf{Category} & \textbf{\% Explain} & \textbf{\% Correct} & \textbf{Ratio} \\
    \hline
    ECO & 7.65 & 57.50 & 0.13 \\
    ECS & 7.38 & 49.11 & 0.15 \\
    IC  & 22.82 & 30.26 & 0.75 \\
    TOP & 18.06 & 35.00 & 0.52 \\
    \hline
  \end{tabular}
  \caption{Proportion of \textit{explain} versus correct \textit{unambiguous} responses for GPT-4o in each category. 
  }
  \label{tab:explain_correct_gpt}
\end{table}

\end{document}